\documentclass[graybox]{svmult}

\usepackage{mathptmx}       
\usepackage{helvet}         
\usepackage{courier}        
\usepackage{type1cm}        
\usepackage{makeidx}         
\usepackage{graphicx}        
\usepackage{multicol}        
\usepackage{multirow}
\usepackage{booktabs}
\usepackage[bottom]{footmisc}
\usepackage{xcolor}
\usepackage[T1]{fontenc}
\usepackage{hyperref}
\usepackage{enumitem}
\usepackage[nolist,nohyperlinks]{acronym}
\begin{acronym}
\acro{HRI}{human-robot interaction}
\acro{AI}{artificial intelligence}
\acro{SAR}{socially assistive robotics}
\acro{HCI}{human-computer interaction}
\acro{ASD}{autism spectrum disorders}
\end{acronym}

\makeindex             

\begin{document}

\title*{An extended framework for characterizing social robots}

\author{Kim Baraka\thanks{Both authors contributed equally to the chapter.}, Patr\'{i}cia Alves-Oliveira\textsuperscript{*}, and Tiago Ribeiro}
\authorrunning{Kim Baraka, Patr\'{i}cia Alves-Oliveira, and Tiago Ribeiro}
\institute{Kim Baraka \at Robotics Institute, Carnegie Mellon University, Pittsburgh, PA 15213, USA 
\at INESC-ID and Instituto Superior T\'{e}cnico, Universidade de Lisboa, 2744-016 Porto Salvo, Portugal
\\ \email{kbaraka@andrew.cmu.edu}
\and Patr\'{i}cia Alves-Oliveira
\at Instituto Universit\'{a}rio de Lisboa (ISCTE-IUL) and CIS-IUL, 1649-026 Lisbon, Portugal
\at INESC-ID, 2744-016 Porto Salvo, Portugal\\ \email{patricia\_alves\_oliveira@iscte-iul.pt}
\and Tiago Ribeiro 
\at INESC-ID and Instituto Superior T\'{e}cnico, Universidade de Lisboa, 2744-016 Porto Salvo, Portugal
\\ \email{tiago.ribeiro@gaips.inesc-id.pt}}
\maketitle

\abstract{Social robots are becoming increasingly diverse in their design, behavior, and usage. In this chapter, we provide a broad-ranging overview of the main characteristics that arise when one considers social robots and their interactions with humans. We specifically contribute a framework for characterizing social robots along $7$ dimensions that we found to be most relevant to their design. These dimensions are: appearance, social capabilities, purpose and application area, relational role, autonomy and intelligence, proximity, and temporal profile. Within each dimension, we account for the variety of social robots through a combination of classifications and/or explanations. Our framework builds on and goes beyond existing frameworks, such as classifications and taxonomies found in the literature. More specifically, it contributes to the unification, clarification, and extension of key concepts, drawing from a rich body of relevant literature. This chapter is meant to serve as a resource for researchers, designers, and developers within and outside the field of social robotics. It is intended to provide them with tools to better understand and position existing social robots, as well as to inform their future design.}

\section{Introduction}
\label{sec:intro}

\subsection{Social humans, social robots}
\label{subsec:motivation}
Humans are inherently social beings, spending a great deal of their time establishing a diverse range of social connections. Their social nature is not only demonstrated by their social behavior~\cite{homans1974social}, but also possesses a biological basis~\cite{frith2010social}. This social dimension prompts human beings to involuntarily ascribe social qualities even to non-human media, such as technological artifacts, often treating them similarly to how they would treat humans or other living beings~\cite{nass1994computers}. This disposition stems from the general human tendency of ascribing human-like qualities to non-human entities, called \textit{anthropomorphism}, which has been observed and demonstrated in several contexts~\cite{epley2007seeing}. These phenomena therefore place technologies capable of social interactions with humans as unique technological innovations. In particular, \textit{social robots}, i.e., robots deliberately designed to interact with humans in a social way, open up a new paradigm for humans to communicate, interact, and relate to robotic technologies.

The integration of a social dimension in the design of robots has generally been following two approaches. First, existing robotic technologies are being enhanced with social capabilities for more fluid interactions with humans. Second, social robots are being developed for new application areas where the social dimension is central, and beyond a mere interface. As a result of these approaches, social robots have been deployed in a wide variety of contexts, such as healthcare~\cite{broadbent2009acceptance}, education~\cite{belpaeme2018social}, companionship~\cite{dautenhahn2005robot}, and others (refer to Section~\ref{subsec:purpose} for a discussion of application areas). They offer a spectrum of interactions that is being continuously enriched by researchers from a variety of disciplines. The field of \ac{HRI}, as an expanding field of research, reflects this observation.
 
 \ac{HRI} is a multidisciplinary field bringing together researchers from an eclectic set of disciplines, including robotics, computer science, engineering, \ac{AI}, machine learning, \ac{HCI}, design, art, animation, cognitive science, psychology, sociology, ethology, and anthropology~\cite{fong2002survey,murphy2010human, baxter2016characterising, alves2016psychological, eyssel2017experimental}. The multidisciplinarity inherent to this field of research provides contributions and advancements nurtured by scholars from different backgrounds in the conception, design, and implementation of social robots. 
 In addition to development, \ac{HRI} aims to evaluate how well such robots perform or serve the purpose they were designed for, being concerned with proper evaluation, testing, and refinement of these technologies. The result is a rich multidisciplinary effort to create engaging robots that can sustain personalized interactions with humans, adapt to the task at hand and to the interaction flow, but also understand and model aspects pertaining to the humans, such as affect and cognition~\cite{ho2010modelling, leite2013social}.

In this chapter, we provide a framework for characterizing social robots that encompasses major aspects to consider when designing them and their interactions with humans. Our framework is focused on interactive robots that possess a social component in their design. Specifically, we use the term ``social robots'' to denote ``socially interactive robots'' as defined by Fong et al.~\cite{fong2002survey}, namely robots  that have one or more of the following abilities: (1) communicating using natural language or non-verbal modalities (such as lights, movements, or sound), (2) expressing affective behaviors and/or perceiving human emotions, (3) possessing a distinctive personality or character, (4) modeling social aspects of humans, (5) learning and/or developing social skills, and (6) establishing and maintaining social relationships~\cite{fong2002survey}.

\begin{figure*}[t]
    \centering
    \includegraphics[width=1.00\textwidth]{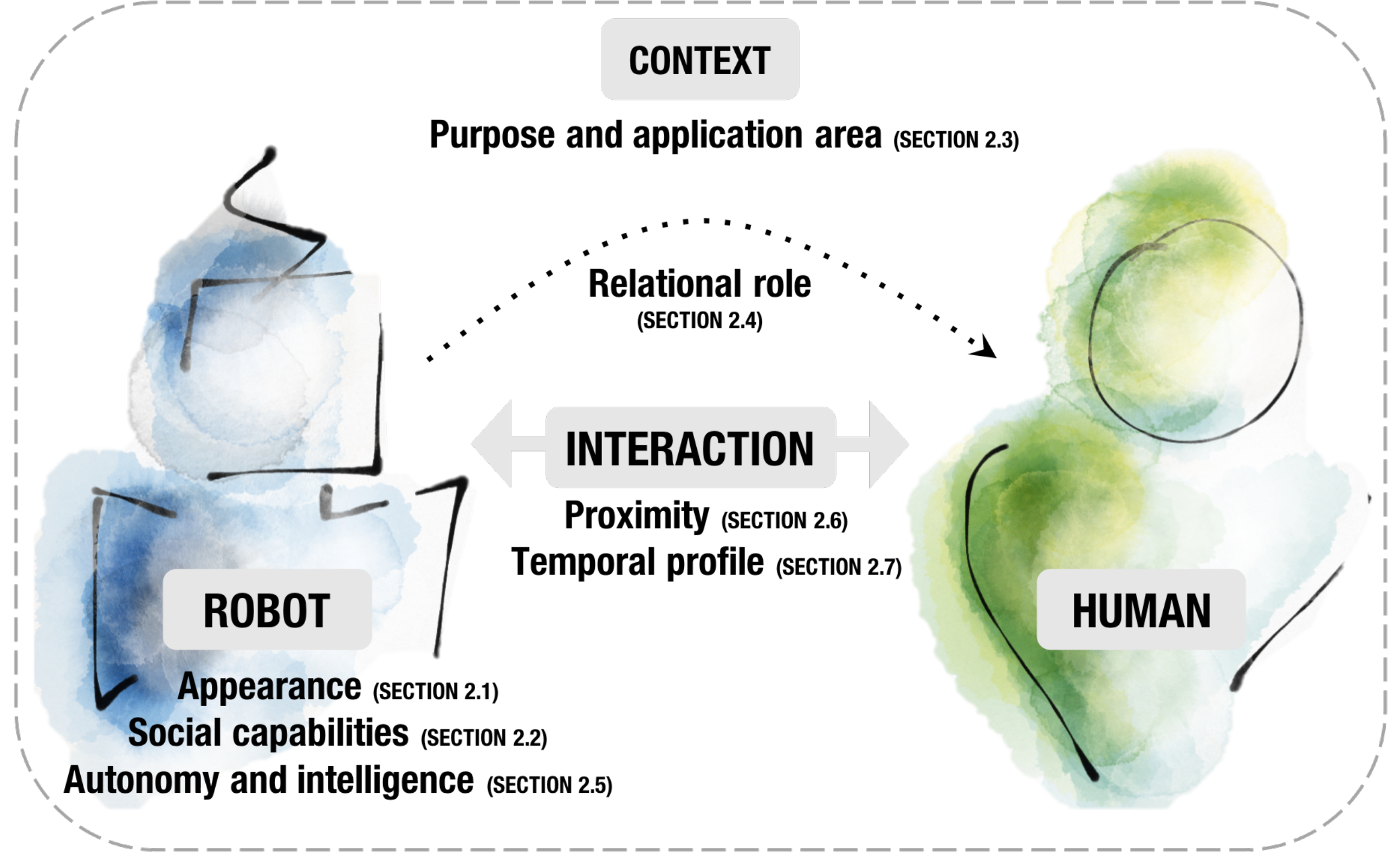}
    \caption{Visual summary of the $7$ dimensions of our framework, positioned in relation to the robot, the interaction, and the context. Each dimension will be further broken down and discussed separately in Section~\ref{sec:framework}.}
    \label{fig:intro}
\end{figure*}

Our framework builds upon existing work within the field of \ac{HRI}, providing a holistic understanding about the state of the art, while aiming at unifying, clarifying, and extending key concepts to be considered in the design of social robots. Specifically, our framework comprises several dimensions we identified to be of major relevance to the design of social robots. We summarize the $7$ dimensions considered in Figure~\ref{fig:intro}. Some of these dimensions relate to the robot itself -- namely \textit{appearance}, \textit{social capabilities}, and \textit{autonomy/intelligence} --, others relate to the interaction -- namely \textit{proximity} and \textit{temporal profile} --, and the remaining ones relate to the context -- namely robot \textit{relational role} and \textit{purpose / application area}. We envision this framework to be used broadly in order to gain a better understanding of existing social robots, as well as to inform the design and development of future ones.

\subsection{Brief summary of frameworks for characterizing social robots}
\label{subsec:existing}
Before outlining the content of our framework, it is useful to first look at existing frameworks for classifying social robots. In particular, existing taxonomies, as such from Fong et al.~\cite{fong2002survey}, Yanco et al.~\cite{yanco2004classifying}, Shibata~\cite{shibata2004overview}, and Dautenhahn~\cite{dautenhahn2007socially}, are useful to get an understanding of different aspects that may be included in the design space of social robots in \ac{HRI} research. While this list of frameworks is not exhaustive, we chose these particular ones to base our framework on, as they provide a broad range of classifications and definitions that relates to the scope of this chapter.

As such, Fong et al.~\cite{fong2002survey} contributed a taxonomy of design methods and system components used to build socially interactive robots. These components include robot social capabilities, several design characteristics, and application domains.

Additionally, Yanco et al.~\cite{yanco2004classifying} provided a framework that included elements of social robot's design, such as the role that a robot can have when interacting with humans, the types of tasks that robots can perform, different types of robot morphology, and the level of autonomy at which robots can operate.

Similarly, Shibata~\cite{shibata2004overview} provided a taxonomy for the function and purpose of social robots by considering different ways of using them for psychological enrichment. Therefore, Shibata classified human-robot interactions in terms of the duration of these interactions and in terms of design characteristics (e.g., robot's appearance, hardware, and software functionalities), accounting for culture-sensitive aspects.

Moreover, Dautenhahn~\cite{dautenhahn2007socially} focused on different evaluation criteria to identify requirements on social skills for robots in different application domains. The author identified four criteria, including contact between the robot and the human (which can vary from no contact or remote contact to repeated long-term contact), the extent of the robot's functionalities (which can vary from limited to a robot that learns and adapts), the role of the robot (which can vary between machine or tool to assistant, companion, or partner), and the requirement of social skills that a robot needs to have in a given application domain (which can vary from not required/desirable to essential). The author further explains that each evaluation criteria should be considered on a continuous scale.

Taken together, these classifications and taxonomies have gathered essential aspects for the characterization and design of social robots. Despite each of them being unique in its contribution, we can see the existence of some overlapping terms and ideas between them. We now discuss our extended framework in the next section.

\subsection{Overview of our extended framework}
\label{subsec:overview}

Our framework leverages the existing ones discussed previously as a starting point and goes beyond the individual frameworks discussed. In particular, it focuses on the following features:

\begin{itemize}
    \item \textbf{Unification ---} The existence of multiple available perspectives in \ac{HRI} often results in scattered concepts and classifications. In this chapter, we aim at merging aspects of the literature on social robots and related fields in a self-contained and consistent resource. 
   \item \textbf{Breadth ---} Existing individual taxonomies often focus on specific aspects relevant to the main line of research of their authors, and may not provide satisfactory coverage. Our framework includes dimensions related to the design of the robot itself, but also of the interaction and context.
    \item \textbf{Recency ---} In recent years, we have observed some important developments in robotic technologies, which have taken robots outside of research laboratory settings and enabled them to be deployed ``in the wild''. We incorporate those recent developments in our work.
    \item \textbf{Clarity ---} Concepts associated with \ac{HRI} are often difficult to define, and as a result clear definitions may not always be available. This lack of clarity may impede communication within the field, or result in inconsistent concepts. In this chapter, we attempt to clarify some important key concepts, such as the distinction between embodiment and purpose, or the concepts of autonomy and intelligence for social robots.
\end{itemize}

With these points in mind, we list below our focuses within each of the $7$ dimensions considered.

\begin{enumerate}
    \item \textbf{Appearance ---} We present a broad classification of robot appearances, synthesizing and going beyond existing ones (Section~\ref{subsec:appearance}).
    \item \textbf{Social capabilities ---} We contribute a repositioning of existing classifications aiming to clarify how existing categories related to each other (Section~\ref{subsec:social}).
    \item \textbf{Purpose and application area ---} We discuss a cross-section of purposes for social robots, and benefiting application areas, with selected examples that include recent developments in the field (Section~\ref{subsec:purpose}). 
    \item \textbf{Relational role ---} We provide a straightforward and broad classification of the robot's role in relation to the human(s) (Section~\ref{subsec:role}). 
    \item \textbf{Autonomy and intelligence ---} We clarify the related but distinct concepts of autonomy and intelligence, and discuss their quantification (Section~\ref{subsec:autonomy}).
    \item \textbf{Proximity ---} We classify interactions according to their spatial features (Section~\ref{subsec:proximity}).
    \item \textbf{Temporal profile ---} We look at several time-related aspects of the interaction, namely timespan, duration, and frequency (Section~\ref{subsec:temporal}). 
\end{enumerate}

It is to be noted that our framework is not meant to be exhaustive, but rather to provide the reader with major aspects that shape social robots and their interactions with humans. While our focus in illustrating the presented concepts will be on single human - single robot interactions, the concepts may also apply for group interactions involving more than one robot and/or more than one human. Additionally, even though this framework was developed with social robots in mind, some dimensions may also be of relevance to robots without a social component in their design, such as for example in the ``appearance'' dimension. In the following section, we delve into each of the $7$ dimensions of our framework. We then end this chapter with a brief discussion on designing social robots within the resulting design space.

\section{Framework description}
\label{sec:framework}

We now provide a description of each of the $7$ dimensions of our framework. The dimensions purposefully operate at different levels, according to the aspects that are most relevant to the design of social robots. In some dimensions, we provide a classification into different categories and possibly subcategories (namely Sections \ref{subsec:appearance}, \ref{subsec:purpose}, \ref{subsec:role}, \ref{subsec:proximity}, and \ref{subsec:temporal}). In others, we focus on clarifying or reinterpreting existing distinctions in categories or scales (namely Sections \ref{subsec:social} and \ref{subsec:autonomy}). Due to different levels of research and relevant content in each, some dimensions are addressed in more depth than others. Also, since the discussions of dimensions are not dependent on each other, we invite the reader to jump to their subsections of interest.

\subsection{Appearance}
\label{subsec:appearance}
The mere physical presence of robots in a shared time and space with humans sparks crucial aspects of a social interaction. Indeed, \textit{embodiment}, a term used to refer to the idea that ``intelligence cannot merely exist in the form of an abstract algorithm but requires a physical instantiation, a body''~\cite{pfeifer2001understanding}, plays an important role in the perception and experience of interacting with intelligent technology. Indeed, literature supports that physical embodiment influences the interaction between humans and robots~\cite{lee2006physically, wainer2007embodiment, powers2007comparing, mumm2011designing, fasola2011comparing, fasola2011comparing, li2015benefit, kennedy2015comparing}. In particular, the physical appearance of a robot \textit{per se}, was shown to have a strong influence on people regarding aspects like perception, expectations, trust, engagement, motivation and usability~\cite{jordan1998human, disalvo2003seduction, breazeal2004designing}.

Several taxonomies were developed in order to create representative classifications for a robot's appearance. To cite a few, Shibata~\cite{shibata2004overview} classified robots as being human type, familiar animal type, unfamiliar animal type, or imaginary animals / new character type. Additionally, Fong et al.~\cite{fong2002survey} considered anthropomorphic, zoomorphic, caricatured, and functional categories. The amount of classifications present in the literature urges for a unified and broad classification for social robot appearances. Building upon the existing classifications, we introduce a broad classification that encompasses main categories described by other authors, as well as new categories and subcategories. Our classification targets only and exclusively a robot's \textit{physical appearance}, as distinct from any type of robot behavior, i.e., ``robot at rest''.

\begin{figure*}
    \centering
    \includegraphics[width=.80\textwidth]{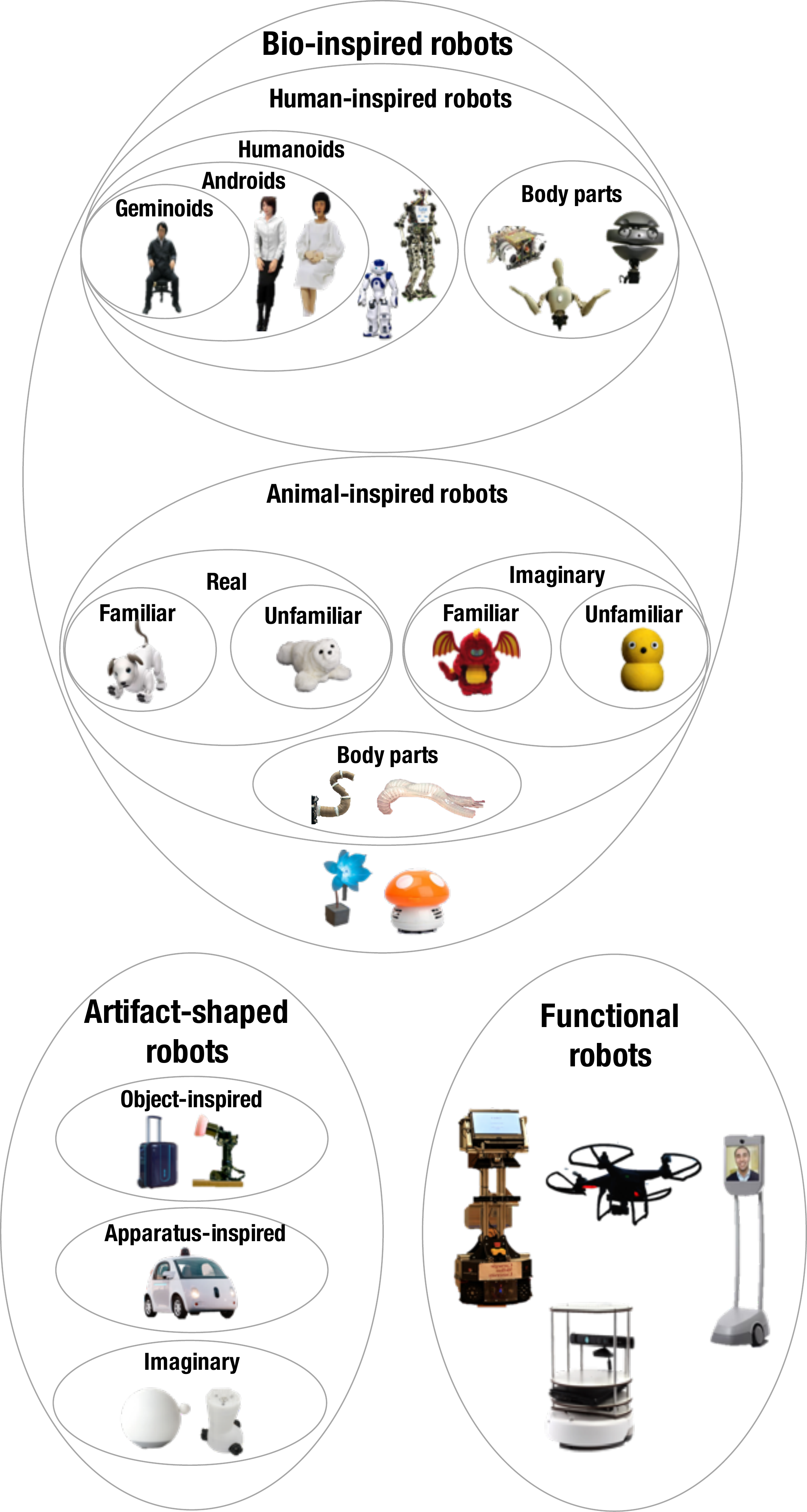}
    \caption{Summary of our robot appearance classification. This classification was based on prior work from Fong et al.~\cite{fong2002survey} and Shibata~\cite{shibata2004overview}, and was unified, extended, elaborated, and clarified in the present chapter. Although the focus is on social robots, its scope is general enough to encompass appearances of robots without a social component in their design. \textbf{List of robots shown (left-to-right, top-to-bottom)} \textit{Bio-inspired robots:} HI-4, ERICA, Kodomoroid, NAO, LOLA, Robotic Eyes, Elumotion, EMYS, AIBO, PARO, DragonBot, Keepon, GoQBot, Meshworm, Robotic Flower, Lollipop Mushroom. \textit{Artifact-shaped robots:} Travelmate, AUR, Google self-driving car, Greeting Machine, YOLO. \textit{Functional robots:} CoBot, Quadcopter, Beam, TurtleBot.}
    \label{fig:appearance}
\end{figure*}

We contribute to the study of social robot's appearance in the following ways: (1) we integrate similar terms already present in the robot appearance classification literature, (2) we add new terms to existing classifications as they were not represented in the literature but urged for a classification, and (3) we attempt to clarify concepts related to different categories. Our unified classification is visually represented in Figure~\ref{fig:appearance}. We considered the following categories of robot appearances: \textit{bio-inspired}, including \textit{human-inspired} and \textit{animal-inspired}, \textit{artifact-shaped}, and \textit{functional}, each with several further subcategories (see Figure~\ref{fig:appearance}). We generated this classification with a holistic mindset, meaning it can serve to classify existing robots, but also to inform the design of future ones. Although devised with social robots in mind, it is general enough to be applied to any robot, independent of its social capabilities. We now provide a description of each category in our classification.

\begin{enumerate}
\item \textbf{Bio-inspired ---} Robots in this category are designed after biological organisms or systems. This includes human-inspired and animal-inspired robots (described next), as well as other bio-inspired robots, such as robotic plants (e.g., the robotic flower\footnote{\href{http://www.roboticgizmos.com/android-things-robotic-flower/}{http://www.roboticgizmos.com/android-things-robotic-flower/}}) and fungi (e.g., the Lollipop Mushroom robot\footnote{\href{https://www.amazon.com/Lollipop-Cleaner-Mushroom-Portable-Sweeper/dp/B01LXCBM3E}{https://www.amazon.com/Lollipop-Cleaner-Mushroom-Portable-Sweeper/dp/B01LXCBM3E}}).
\begin{enumerate}
\item \textbf{Human-inspired ---}
	Robots in this category are inspired by features of the human body, including structure, shape, skin, and facial attributes. Human-inspired robots not only include full-body designs, but also robots designed after human body parts. When designed after the full-human body, they are called \textit{humanoids}. The level of fidelity can vary from a highly mechanical appearance, such as the LOLA robot~\cite{buschmann2009humanoid}, to a highly human-like appearance that includes skins and clothes, such as the ERICA robot~\cite{glas2016erica}, or even include an intermediate between this two, in the case of the NAO robot\footnote{\href{https://www.softbankrobotics.com/emea/en/nao}{https://www.softbankrobotics.com/emea/en/nao}}. 
	For humanoids, it is worth mentioning the case in which they strongly resemble the human outer appearance and are covered with flesh- or skin- like materials, in which case they are often referred to as \textit{androids} (if they possess male physical features) or \textit{gynoids} (if they possess female physical features). An example of a gynoid is the Kodomoroid robot\footnote{\href{http://www.geminoid.jp/en/robots.html}{ http://www.geminoid.jp/en/robots.html}}. Additionally, a special case of androids/gynoids are \textit{geminoids}, which are designed after an existing human individual -- i.e., it is a ``robotic twin'' -- such as Geminoid HI-4\footnote{\href{http://www.geminoid.jp/projects/kibans/resources.html}{http://www.geminoid.jp/projects/kibans/resources.html}}, the tele-operated robotic twin of Hiroshi Ishiguro.	
	On the other hand, some robots are inspired by individual \textit{parts of the human body}. These include robotic arms, e.g., Elumotion Humanoid Robotic Arm\footnote{\href{http://elumotion.com/index.php/portfolio/project-title-1}{http://elumotion.com/index.php/portfolio/project-title-1}}, robotic hands~\cite{liu2007modular}, robotic heads such as the EMYS robot~\cite{kkedzierski2013emys}, robotic torsos,~\cite{shidujaman2018roboquin}, and robotic facial features, such as robotic eyes~\cite{cannata2006design},.
	It is worth mentioning that high-fidelity human-inspired robots are often subject to uncanny valley effects~\cite{mori1970uncanny}. Being highly but not totally human-like, they elicit feelings of eeriness, and hence should be designed bearing these possible effects in mind.\\

\item \textbf{Animal-inspired ---}
	Robots in this category are inspired by animals or by creatures possessing animal traits of appearance. On the one hand, they may be inspired by \textit{real animals}, for which we consider inspiration from \textit{familiar} animals, like the AIBO\footnote{\href{https://us.aibo.com/}{https://us.aibo.com/}} dog-inspired robot, and inspiration from \textit{unfamiliar} animals, such as the PARO\footnote{\href{http://www.parorobots.com/}{http://www.parorobots.com/}} baby seal robot. 
    The distinction between familiar and unfamiliar animals is emphasized by Shibata~\cite{shibata2004overview}. According to the author, familiar animals are those whose behavior can be easily recognized, such as pets; while unfamiliar animals are those that most people know something about but are not totally familiar with, and have rarely interacted with them before, such as savanna animals. The same author mentioned that when robots are designed to resemble an unfamiliar animal they can be more easily accepted due to the lack of exposure to their typical behavior. It is documented in the literature that people hold strong expectations when faced with the possibility of interacting with a social robot~\cite{spence2014welcoming}, wherein robots whose embodiment matches its abilities are perceived more positively~\cite{goetz2003matching, li2010cross, komatsu2012does}. However, it is to be noted that familiarity is a subjective concept depending on culture and individual experiences, making this distinction flexible. On the other hand, animal-inspired robots can also be \textit{imaginary}, meaning they possess animal-like features but are not designed after a real animal. They can either be \textit{familiar}, i.e., designed after familiar imaginary animals ``existing'' in fantasy worlds, like cartoon characters or legendary creatures (e.g., DragonBot~\cite{short2014train}), or \textit{unfamiliar}, i.e., robots that are purely created from imagination, such as Miro\footnote{\href{http://consequentialrobotics.com/miro/}{http://consequentialrobotics.com/miro/}} and Keepon\footnote{\href{https://beatbots.net/my-keepon}{https://beatbots.net/my-keepon}}. In addition, this category includes robots designed after \textit{animal body parts}, such as the GoQBot designed as a caterpillar part~\cite{lin2011goqbot}, the Meshworm designed after the oligochaeta~\cite{seok2010peristaltic}, and robotic soft tentacles~\cite{jorgensen2018interaction}.
\end{enumerate}
\item \textbf{Artifact-shaped ---}
    Robots in this category bear the appearance of physical human creations or inventions. They may be inspired by \textit{objects}, such as furniture and everyday objects, e.g., the AUR robotic desk lamp~\cite{hoffman2010effects}, the Mechanical Ottoman robotic footstool~\cite{sirkin2015mechanical}, and the Travelmate robotic suitcase\footnote{\href{https://travelmaterobotics.com/}{https://travelmaterobotics.com/}}. They may also be inspired by an existing \textit{apparatus}, demonstrating how existing apparatuses can become robotic systems while maintaining the same appearance, such as self-driving cars (e.g., the Google self-driving car\footnote{\href{https://waymo.com/}{https://waymo.com/}}), but also everyday apparatuses like toasters, washing machine, etc. Additionally, artifact-shaped robot may be \textit{imaginary}, i.e., translating the invention of the designer, such as the Greeting Machine robot~\cite{anderson2018greeting} or YOLO~\cite{alves2017yolo, alves2018yolo}.
\item \textbf{Functional ---} The appearance of robots included in this category is merely the sum of appearances of the technological pieces needed to achieve a given task or function. This means that their appearance leans more towards mechanical aspects. Examples are quadcopters, or mobile robots such as the CoBots~\cite{veloso2015cobots}, the TurtleBot\footnote{\href{https://www.turtlebot.com/}{https://www.turtlebot.com/}}, and the Beam\footnote{\href{https://suitabletech.com/}{https://suitabletech.com/}}.
\end{enumerate}

As a side note, shape-shifting robots, modular robots, or polymorphic robots~\cite{balch2002robot, yim2002modular, yim2007modular, li2009amoeba} are all examples of hybrid robots that can fit into more than one category depending on their configuration. Also, robotic swarms are examples of multi-robot systems that may be perceived as a single entity, i.e., more than the sum of individual robots (homogeneous or heterogeneous)~\cite{kolling2016human}, however they are they are not part of our classification, because they are too dependent on the configuration and behavior of the swarm. Moreover, the actual process of assigning categories to existing robots always carries a certain degree of subjectivity, which relates to different possible perceptions of the same robot appearance, depending or not on the context, the behavior of the robot, and so on. The clearest example in our classification would be the distinction between familiar and unfamiliar, which strongly depends on people's cultural background and personal experiences. Those differences in perception should be accounted for when designing robot appearances.

Our presented classification is not intended to offer a clear-cut or rigid boundary between categories of robots. Rather, it represents a useful guideline for categorizing robots based on major distinguishing features. It does encourage the view of robot design as a spectrum, providing fluidity to their design and allowing for the combination of different elements of our classification.

A robot's appearance is the most obvious and unique visual attribute, which contributes highly to the interaction~\cite{fink2012anthropomorphism}. Nonetheless, in addition to appearance, there are several factors related to embodiment, such as size, weight, noise, material texture, among others~\cite{disalvo2002all} that may contribute to the perception of the robot during an interaction. More research is needed in order to develop classifications that account for the other factors mentioned above.

\subsection{Social capabilities}
\label{subsec:social}

Social robots vary greatly in their social capabilities, i.e., how they can engage in and maintain social interactions of varying complexities. As such, researchers have classified and defined them according to those social capabilities. Based on the work of Fong et al.~\cite{fong2002survey}, we list the different components of a social robot's capabilities as follows:

\begin{itemize}
    \item \textbf{Communicating using natural language or non-verbal modalities ---} Examples of these ways of communication are natural speech \cite{williams2018thank}, motion~\cite{knight2011eight,dragan2013legibility} -- possibly including gaze~\cite{admoni2017social}, gestures or facial expressions --, lights~\cite{baraka2018mobile,szafir2015communicating}, sounds~\cite{bethel2006auditory}, or a combination of them \cite{loffler2018multimodal}. Mavridis~\cite{mavridis2015review} provided a review on verbal and non-verbal interactive communication between humans and robots, defining different types of existing communications such as interaction grounding, affective communications, speech for purpose and planning, among others.
    \item \textbf{Expressing affect and/or perceiving human emotions ---} Beyond Ekamn's five basic emotions~\cite{ekman1992argument} -- anger, disgust, fear, happiness, sadness, and surprise --, this may include more complex affective responses such as empathy. For example, Paiva et al.~\cite{paiva2017empathy} analyzed different ways by which robots and other artificial agents can simulate and trigger empathy in their interactions with humans.
    \item \textbf{Exhibiting distinctive personality and character traits ---} The major components to be considered, according to Robert~\cite{robert2018personality}, are human personality when interacting with a robot, robot personality when interacting with humans, dissimilarities or complementary in human-robot personalities, and aspects that facilitate robot personality. Some companies such as Misty Robotics \footnote{\href{https://www.mistyrobotics.com/}{https://www.mistyrobotics.com/}} are prioritizing the user personalization of a robot's personality as an important feature for future commercial social robots.
    \item \textbf{Modeling and recognizing social aspects of humans ---} Modeling human agents allows for robots to interpret aspects of human behavior or communication and appropriately respond to them. Rossi et al.~\cite{rossi2017user} provide a survey of sample works aimed at profiling users according to different types of features. More advanced models may have to consider theory of mind approaches~\cite{scassellati2002theory}. 
    \item \textbf{Learning and developing new social skills and competencies ---} In addition to being programmed to have social skills, social robots may have the ability to refine those skills with time through adaptation, or even developing new skills altogether. An active area of research that looks at such paradigms is the area of developmental robotics~\cite{lungarella2003beyond}.
    \item \textbf{Establishing and maintaining social relationships ---} Relationships operate over a timespan that goes beyond a few interactions. A number of questions arise when one considers long-term interactions of robots with humans and what it means for a robot to proactively establish and maintain a relationship that is two-sided. Leite et al.~\cite{leite2013social} established some initial guidelines for the design of social robots for long-term interaction. These include  continuity and incremental robot behaviors (e.g., recalling previous activities and self-disclosure), affective interactions and empathy (e.g., displaying contextualized affective reactions), and memory and adaptation (e.g., identifying new and repeated users).
\end{itemize}

Complementary to these components, Breazeal~\cite{breazeal2003toward} distinguished $4$ categories of robot social capabilities: (1) \textit{socially evocative}, denoting robots that were designed mainly to evoke social and emotional responses in humans, leveraging the human tendency to anthropomorphize~\cite{epley2007seeing}. Therefore, despite expected social responsiveness, the robot's behavior does not necessarily reciprocate; (2) \textit{social interface}, denoting robots that provide a ``natural'' interface by using human-like social cues and communication modalities. In this sense, the social behavior of humans is only modeled at the interface level, which normally results in shallow models of social cognition in the robot; (3) \textit{socially receptive}, denoting robots that are socially passive but that can benefit from interaction. This category of robots is more aware of human behavior, allowing humans to shape the behavior of the robot using different modalities, such as learning by demonstration. Also, these robots are socially passive, responding to humans' efforts without being socially pro-active; and (4) \textit{sociable}, denoting robots that pro-actively engage with humans, having their own internal goals and needs in order to satisfy internal social aims (drives, emotions, etc.). These robots require deep models of social cognition not only in terms of perception but also of human modelling.

In addition to this list, Fong et al.~\cite{fong2002survey} added the following $3$ categories: (5) \textit{socially situated}, denoting robots that are surrounded by a social environment that they can perceive and react to. These robots must be able to distinguish between other social agents and different objects that exist in the environment; (6) \textit{socially embedded}, denoting robots that are situated in a social environment and interact with other artificial agents and humans. Additionally, these robots can be structurally coupled with their social environment, and have partial awareness of human interactional structures, such as the ability to perform turn-taking; and (7) \textit{socially intelligent}, including robots that present aspects of human-style social intelligence, which is based on deep models of human cognition and social competence.

Although robots have been classified according to their different social capabilities, it is yet unclear how these categories relate to each other. Are they part of a spectrum? Are they separate categories altogether? We argue that evaluating social capabilities of robots can be understood according to two main dimensions:

\begin{enumerate}
\item\textbf{The depth of the robot's actual social cognition mechanisms.}
\item \textbf{The human perception of the robot's social aptitude.}
\end{enumerate}

\begin{figure}
\sidecaption[t]
\includegraphics[width=7.5cm]{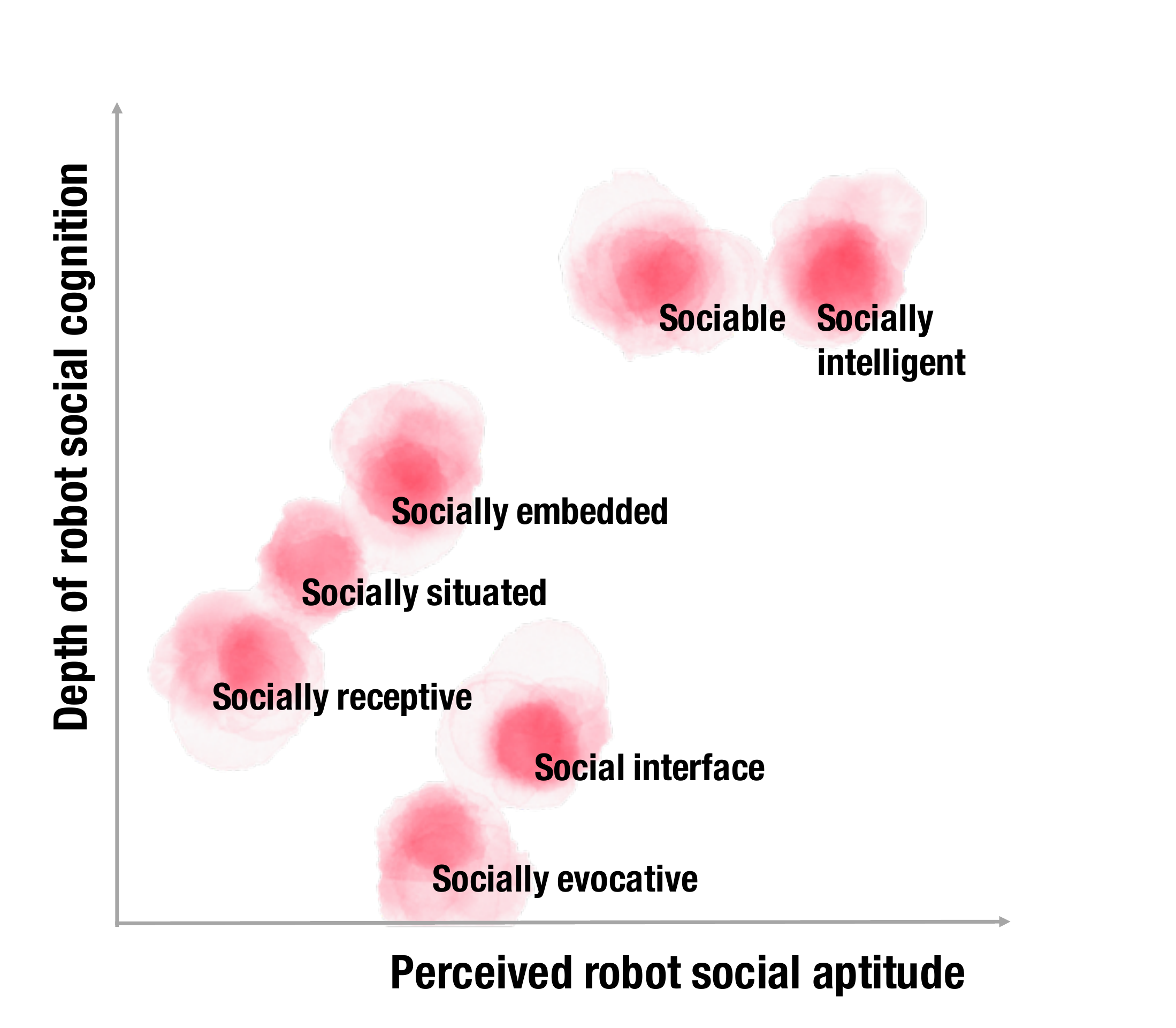}
\caption{Positioning of the classifications of Breazeal  \cite{breazeal2003toward} and Fong et al. \cite{fong2002survey} according to our proposed two-dimensional space formed by (1) the depth of the robot's social cognition mechanisms, and (2) the expected human-perceived level of robot social aptitude. This figure is merely illustrative and color patches deliberately fuzzy, as we do not pretend to have the tools to actually quantify these dimensions according to any scale.}
\label{fig:social_capability} 
\end{figure}

Given these dimensions, and in light of the existing categories presented above, we propose a two-dimensional space map, providing a clearer understanding of the social capabilities of robots. This map is presented in Figure~\ref{fig:social_capability} for illustrative purposes. As it can be seen in the figure, socially evocative robots have the least depth of social cognition but are perceived as rather socially apt. A social interface typically possesses some additional cognition mechanisms to allow for easy communication with the range of the robot's functionality; it also possibly results in a slightly higher perceived social aptitude thanks to its more versatile nature. Socially receptive robots, socially situated, and socially embedded robots possess increasing depth in their social cognition, and as a result increasing perceived social aptitude. For socially embedded robots, the perceived aptitude may vary according to the degree of awareness about interactional structure the robot has. On the outskirts of our map we find sociable and socially intelligent robots, with much deeper models of social cognition.

\subsection{Purpose and application area}
\label{subsec:purpose}

\begin{figure*}
    \centering
    \includegraphics[width=\textwidth]{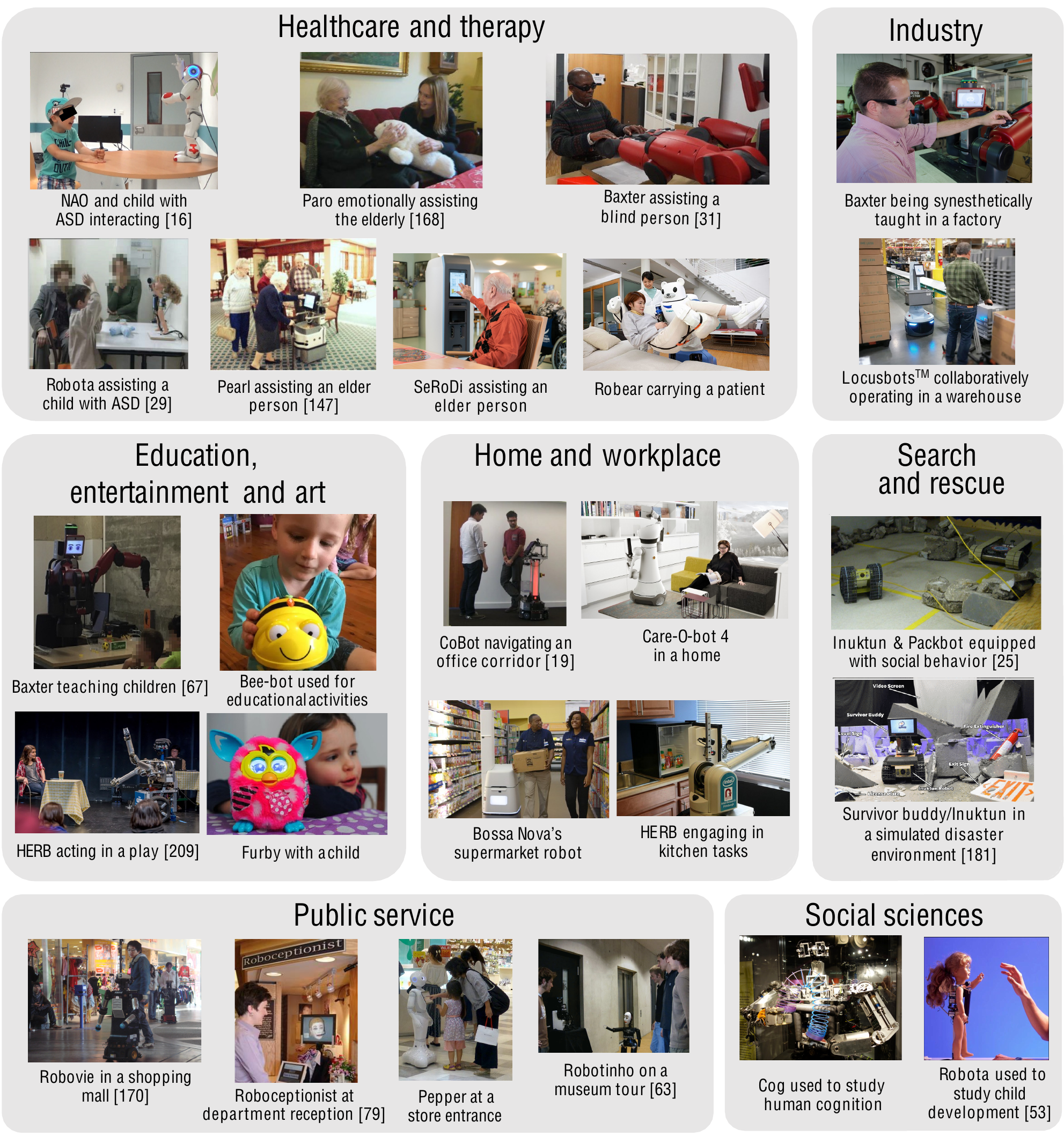}
    \caption[A cross-section of main application areas for social robots with selected examples, and emphasis on the possibility of more than one purpose for the same physical robot.]{A cross-section of main application areas for social robots with selected examples, and emphasis on the possibility of more than one purpose for the same physical robot, e.g., Baxter appears in healthcare, industry, and education. Education and entertainment/art were merged for conciseness. All images were reproduced with permission of the authors, companies or copyright owners. Additional credits, when applicable, are included in a footnote\footnotemark.}
    \label{fig:applications}
\end{figure*}

In this section, we discuss social robots according to their purpose, i.e., what types of goals they are designed to achieve, as well as benefiting application areas. Figure~\ref{fig:applications} summarizes the main purposes and application areas included in this section, with illustrative examples.

\nocite{shibata2011robot}
\nocite{baraka2019optimization}
\nocite{pollack2002pearl}
\nocite{bethel2009robots}
\nocite{srinivasan2012social}

\subsubsection*{A note on purpose as being distinct from embodiment}
In traditional engineering practice, the physical characteristics of a technological device (e.g., toaster, microwave, typewriter, manufacturing machine) tend to be strongly coupled with its purpose, i.e., the task it was designed to achieve. With the advent of personal computers and smartphones, we moved away from defining those devices solely by their purpose. For instance, it would be inappropriate to call a modern computer an ``electronic typewriter'' or even a smartphone an ``electronic phone'', because those devices can serve an immense variety of uses, thanks to software applications that constantly create new purposes for them. Similarly, even though some robots may currently be designed for a specific purpose in mind, some robots may possess a set of skills that can prove useful in a variety of scenarios, sometimes across completely different application areas. As a result, (1) many different robots can be programmed to be used for the same purpose, but also (2) a single robot can be used for many different purposes. For example, a robot such as NAO has been used across a large variety of purposes, both in research and industry, from playing soccer~\cite{graf2009robust} to assisting individuals with cognitive impairments~\cite{shamsuddin2012initial,esteban2017build} or teaching children~\cite{yadollahi2018deictic,alves2019empathic}.

\footnotetext{\scriptsize{\textbf{Additional credits (left-to-right, top-to-bottom)} \textit{Paro}: Credits AIST, Japan; \textit{Baxter (industry)}: Courtesy of Rodney Brooks; \textit{SeRoDi}: Source Fraunhofer IPA, Photographer Rainer Bez (2015); \textit{Robear}: Credits RIKEN; \textit{Bee-bot}: Credits Ben Newsome, Fizzics Education; \textit{Care-O-bot}: Source Phoenix Design (2015); \textit{Furby}: Credits Robert Perry; \textit{HERB}: Courtesy of Siddhartha Srinivasa; \textit{Robovie}: Courtesy of Masahiro Shiomi; \textit{Pepper}: Retrieved from Wikimedia Commons under the \href{https://en.wikipedia.org/wiki/GNU_Free_Documentation_License}{GNU Free Documentation License}, Author Nesnad; \textit{Robotinho}: Credits University of Freiburg; \textit{Robota (social sciences)}: retrieved from~\cite{billard2007building}.}}

There remains, however, a general tendency to define robots by characteristics of their programmed behavior, which can be limiting or inappropriate. As an example, we see locutions of the form ``educational robots'', ``therapeutic robots'', ``pet robots'', and so on. The Baxter robot\footnote{\href{https://www.rethinkrobotics.com/baxter/}{https://www.rethinkrobotics.com/baxter/}}, for instance, is often referred to as a ``collaborative industrial robot'' (or co-bot), because it has been used quite often in such a setting. However, it has also been used in very different applications, such as assistance for the blind~\cite{bonani2018my}, or education~\cite{fernandez2018may}, and hence the naming is reductive. Similarly, a ``pet robot'' such as the AIBO dog-inspired robot has been used in contexts where it is far from being considered a pet, such as playing soccer with other robots~\cite{stone2007intelligent}.

Of course, the embodiment of the robot may restrict its capabilities and hence the type of tasks it may be able to physically achieve. Also, the robot's hardware may be optimized for a specific interactive application (e.g., Baxter has compliant joints for safer collaboration). Moreover, a robot's appearance, which goes beyond its hardware specifications, may be optimized for human perceptions such as acceptability, likeability, trust, and so on, for a specific intended purpose. However, given the considerations above, we believe that robots should not be defined solely by their purpose, the same way humans are (hopefully) not defined by their profession. As a result, we personally prefer a slightly different language to characterize robots according to their purpose(s): ``robots \textit{for} education'' instead of ``educational robots'', ``robots \textit{for} therapy'' instead of ``therapeutic robots'', and so on. Using this slightly modified language, we now discuss the main purposes and application areas that are benefiting from the use of social robots. In light of our discussion, the presented list is not meant to be selective, as the same robot may be used for more than one purpose.

\subsubsection{Robots for healthcare and therapy} 
\label{subsubsec:healthcare}

Robots are being introduced in the health sector to assist patients and providers in hospitals, at home, or in therapy settings. The type of assistance the robot provides can be generally categorized into physical and/or social. Physically assistive applications include helping patients with reduced mobility or dexterity, such as the elderly~\cite{forlizzi2004assistive} or people with physical impairments~\cite{burgar2000development}. These robots can help to carry out daily tasks, like getting out of bed, manipulating objects, eating, and so on, which can give them a higher sense of autonomy and dignity~\cite{sharkey2012granny}. They may also help in therapy to assist patients in regaining lost physical skills or build new ones~\cite{burgar2000development}. On the other hand, \ac{SAR} focus on providing assistance primarily through social interactions. Feil-Seifer et al.~\cite{feil2005defining} identified a number of applications where \ac{SAR} may have a strong impact, namely in therapy for individuals with cognitive disorders~\cite{scassellati2012robots,cabibihan2013robots}, companionship to the elderly and individuals with neurological disorders or in convalescent care~\cite{burton2013dolphins}, and students in special education. We also believe that robots in the healthcare domain may be used to benefit healthcare providers directly, for example training therapists through robotic simulation of interactions with patients~\cite{baraka2019interactive}.

\subsubsection{Robots for education}
Robots in education are mainly used with children~\cite{kanda2007two,tanaka2007socialization,Kozima08aplayful} because they can increase engagement in learning while favoring an interactive and playful component, which may be lacking in a traditional classroom setting. When designing such educational robots, it is crucial to design for and evaluate long-term interactions, to avoid successes merely due to strong novelty effects~\cite{leite2013social}.

There is a number of formats that educational scenarios can take, where the robot has a different role. Beyond being a teacher delivering material, the robot can also act as a social mediator between children, encouraging dyadic, triadic, and group interactions~\cite{kozima2009keepon}. Moreover, the robot may play the role of a learner in learning-by-teaching scenarios, in which the child teaches the robot and in this process develops their own skills~\cite{jacq2016building}.

\subsubsection{Robots for entertainment and the arts}
The entertainment industry has benefited from the use of robots for their engaging and interactive capabilities. Personal entertainment creations emerged with robotic toys, such as Furby\footnote{\href{https://furby.hasbro.com/en-us}{https://furby.hasbro.com/en-us}} or Bee-Bot\footnote{\href{https://www.bee-bot.us/}{https://www.bee-bot.us/}}, and robotic dolls, such as Hasbro's My Real Baby\footnote{\href{https://babyalive.hasbro.com/}{https://babyalive.hasbro.com/}}. Public entertainment robots have appeared in theme parks and other public entertainment spaces~\cite{madhani2009bringing}. More complex robots with both verbal and non-verbal communication capabilities have been used for more prolonged interaction scenarios such as storytelling~\cite{chen2011survey} or comedy~\cite{bruce2000robot}. Other entertainment applications include interactive shows~\cite{alonso2014human}, acrobatic robots for movie stunts~\cite{pope2018stickman}, and sex robots~\cite{levy2009love}, among others. 

More artistic-oriented applications include robots in the visual arts\footnote{An annual robot art competition is held to encourage the use of robots in the visual arts \href{http://robotart.org/}{http://robotart.org/}}~\cite{pagliarini2009development} and installation art~\cite{augugliaro2014flight}. Social robots have also been deployed in fields of performative arts such as drama~\cite{zeglin2014herb} or dance~\cite{sum2017robot,cappo2018online}, where their embodied intelligence in real-time contexts and their interactivity remain a challenging and rich research challenge. Generally, the inclusion of intelligent robots in the arts and the broader field of computational creativity~\cite{colton2012computational} are questioning definitions and criteria of art, authorship, and creativity.

\subsubsection{Robots for industry}
As industrial robots are becoming more intelligent, they are being equipped with interactional capabilities that allow them to collaborate with humans, mainly in tasks involving manipulation skills. Schou et al.~\cite{schou2018skill} identified several tasks that can benefit from a human-robot collaborative setting, possibly including multi-robot/multi-human teams. These are: logistic tasks (namely transportation and part feeding), assistive tasks (namely machine tending, (pre)assembly, inspection, and process execution), and service tasks (namely maintenance and cleaning). 

Research has shown that robots exhibiting social communication cues in industrial settings are perceived as social entities~\cite{sauppe2015social}. Moreover, Fong et al.~\cite{fong2002survey} emphasized that, in order to achieve true collaboration between humans and robots, the robot must have sufficient introspection to detect its own limitations, must enable bidirectional communication and information exchange, and must be able to adapt to a variety of humans from the novice to the experienced.

\subsubsection{Robots for search and rescue}
Search and rescue is one of the applications in which robots are being investigated as replacements to humans in dangerous environments, such as in natural or human disasters. Even though typical robots in this domain have not been designed with social capabilities, research has shown the importance of ``social intelligence'' in this domain~\cite{fincannon2004evidence}. Bethel et al.~\cite{bethel2008survey} identified the importance of different modalities of social communication in the context of victim approach, across the scale of proxemic zones (i.e., the distancing of the robot to the human), ranging from the public to the personal space. Such modalities include body movement, posture, orientation, color, and sound.

\subsubsection{Robots for assistance in home and workplace}
With the advent of personal robots~\cite{gates2007robot}, the vision is that anyone will have the ability to own and operate a robot, regardless of their skills or experience, thanks to natural and intuitive interfaces \cite{liang2018simultaneous}. Such robots can be deployed in home or workplace environments to assist individuals, reduce their mental and physical load, and increase their comfort and productivity. In the home, personal robots are already cleaning floor surfaces autonomously\footnote{\href{https://www.irobot.com/for-the-home/vacuuming/roomba}{https://www.irobot.com/for-the-home/vacuuming/roomba}}, cooking full meals\footnote{\href{http://www.moley.com/}{http://www.moley.com/}}, and doing laundry\footnote{\href{http://www.laundry-robotics.com/}{http://www.laundry-robotics.com/}}, just to name a few. More ambitious research projects have aimed at designing versatile ``robotic butlers''~\cite{srinivasa2010herb}, that can operate in a variety of tasks across the home.

In the workplace, robots are being used on a daily basis to transport objects, cataloguing inventory, escorting people, delivering messages, among other tasks, in settings such as offices, hospitals\footnote{\href{https://aethon.com/}{https://aethon.com/}}, supermarkets\footnote{\href{http://www.bossanova.com}{http://www.bossanova.com}}, and hotels. The majority of these robots are called service robots and have the capability of navigating in structured indoor environments, mainly corridors as opposed to open public spaces. An example of such service robots is the CoBots~\cite{veloso2015cobots}, developed and deployed at Carnegie Mellon University, servicing multiple floors and having navigated more than $1,000$~km autonomously~\cite{biswas20161}. Other types of robots used in the workplace include tele-presence robots for teleconferencing and virtual visits of remote places~\cite{tsui2011exploring}.

\subsubsection{Robots for public service}
Robots have been deployed in public spaces including malls~\cite{shiomi2009field}, museums~\cite{faber2009humanoid}, exhibition spaces~\cite{jensen2005robots}, and receptions~\cite{gockley2005designing}. Some (but not all) of those robots are mobile, and can navigate in open spaces or in crowds, which makes the design of their behavior challenging and subject to a variety of social constraints~\cite{luber2012socially}. Interactions with such robots have to account for the fact that the robot will interact with a very large number of people, with inevitable differences, and during a short duration. Hence, personalizing the interaction and making it as intuitive as possible (as there is very little adaptation time on the human side) are important design considerations.

\subsubsection{Robots for the social sciences}

Due to the possibility of programming robots to exhibit mechanisms of cognition similar to those of humans, a less publicized purpose of robots is in fields of the social sciences for the study of social development, social interaction, emotion, attachment, and personality~\cite{fong2002survey}. The idea is to use robots as test subjects in controlled laboratory experiments, leveraging the fact that such robots can reproduce consistent behaviors repeatedly and can be controlled to test predictions of human models of cognition. For example, the Cog robot~\cite{scassellati2003investigating} was used to investigate models of human social cognition. Similarly, a doll-like robot, Robota~\cite{billard2007building}, was used in comparative studies for social development theories~\cite{dautenhahn1999studying}. Additionally, robots (human-inspired or other types) can be used as stimuli to elicit behaviors from humans for the development and refinement of theories about human behavior and cognition. For a more detailed discussion on cognitive robotics and its applications outside of technology-related fields, consult Lungarella et al.~\cite{lungarella2003developmental}.

\subsubsection{Other application areas}

The list of application areas and purposes listed above is not comprehensive, but reflects major developments and deployments. To this list we can add: 

\begin{itemize}
\item \textbf{Robots for companionship ---} Dautenhahn~\cite{dautenhahn2004robots} presented a perspective on different possible relationships with personalized (possibly life-long) robotic companions, drawing on literature from human-animal relationships. Situated somewhere between animal pets and lifeless stuffed animals, robotic companions may provide support for socially isolated populations. The technical and design challenges associated with robotic companions are numerous due to the time dimension, and the deployment of robotic pets has raised an ethical concern~\cite{sparrow2002march}. Examples of robotic companions include the Huggable$^{TM}$ robot~\cite{stiehl2009huggable}, the AIBO dog-inspired robot~\cite{friedman2003hardware}, and the Lovot robot\footnote{\href{https://groove-x.com/en/}{https://groove-x.com/en/}}.
\item \textbf{Robots for personal empowerment ---} The ultimate ethically concerned use of robots is to expand human abilities instead of replacing them, and to empower people at an individual level. Examples of personal empowerment that robots may facilitate are physically assistive robots that help people with impairments gain autonomy and dignity, such as prosthetics, exoskeletons, brain-controlled robotic arms~\cite{hochberg2012reach}, and other assistive robots (see Section~\ref{subsubsec:healthcare}). Other examples include robots that are designed to enhance creativity in individuals, such as the YOLO robot~\cite{alves2018yolo}, or tele-presence robots for workers that cannot physically perform the required tasks, such as in the ``Dawn ver. $\mathrm{\beta}$'' cafe in Japan who hired paralyzed people to serve the costumers through a mobile robot controlled by their eye movements\footnote{\href{https://www.bbc.com/news/technology-46466531}{https://www.bbc.com/news/technology-46466531}}.
\item \textbf{Robots for transportation ---} The rise of autonomous driving will revolutionize transportation and the urban environment. Autonomous vehicles (cars, trucks, public transportation, etc.) are expected to operate in environments populated by humans (drivers, pedestrians, bicyclists, etc.), and research is looking at adding social dimensions to their behavior~\cite{nass2005improving,wei2013autonomous, mavrogiannis2019effects}. Additionally, drones will be used in the near future for package delivery\footnote{\href{https://www.amazon.com/Amazon-Prime-Air/b?ie=UTF8\&node=8037720011}{https://www.amazon.com/Amazon-Prime-Air/b?ie=UTF8\&node=8037720011}} and will have to (socially) interact with costumers.
\item \textbf{Robots for space ---} Robots for space exploration are historically known for their low level of interactions with humans. However, as humans are getting more involved in space explorations, social robots are being introduced to assist astronauts in their tasks and daily routines, e.g., the Jet Propulsion Laboratory's Robonaut and Valkyrie~\cite{yamokoski2019robonaut}.
\item \textbf{Robots for technology research ---} Robots can also be used to test theories in fields related to technology, such as testing algorithms and architectures on physical platforms. More generally, robots can provide a platform for developing and testing new ideas, theories, solutions, prototypes, etc., for effective embodied technological solutions and their adoption in society.
\end{itemize}

The application areas mentioned above provide a cross-section of purposes that social robots hold in existing developments and deployments. If we view robots as embodied agents that can carry intelligently complex tasks in the physical and social world, we expect, in the future, to have robots introduced in virtually any application where they can complement, assist, and collaborate with humans in existing roles and expand their capabilities, as well as potentially assume new roles that humans cannot or should not assume.

\subsection{Relational role}
\label{subsec:role}

One of the relevant dimensions that shapes human-robot interaction is the \textit{role} that the robot is designed to fulfill.
The concept of role is an abstract one, for which various different perspectives can be presented. 
In this section, we specifically look at the \textit{relational role} of the robot towards the human.
This is the role that a robot is designed to fulfill within an interaction, and is not necessarily tied to an application area. The relational role the robot has been designed to have is critical to the perception, or even the relationship, that arises between robot and human.
 
Towards clarifying the concept of relational role, it is important to immediately distinguish relational role from role in an activity or application. 
In a specific activity or application, we may expect to find activity-specific roles (as in role-playing), such as teacher, driver, game companion, cook, or therapist. 
These types of roles are defined by the type of activity performed between the robot and humans, therefore making it an open-ended list that is likely to stay in constant evolution, as robots become applied to new fields and tasks.

Given the fuzziness of this concept, there have not been many attempts at generalizing the concept of role of robots within a relation with humans. 
For the rest of this section, we will present and analyze some broader definitions from the existing literature, to conclude by contributing a broad classification that attempts to agglomerate the main concepts of the pre-existing ones while containing and extending them.

Scholtz et al. presented a list of interaction models found in \ac{HRI}~\cite{scholtz2003theory}. They included roles that humans may have towards a robot in any \ac{HRI} application. The list defines the roles of the Supervisor, who monitors and controls the overall system (single or multiple robots), while acting upon the system's goals/intentions; the Operator, who controls the task indirectly, by triggering actions (from a set of pre-approved ones), while determining if these actions are being carried out correctly by the robot(s); the Mechanic, who is called upon to control the task, robot and environment directly, by performing changes to the actual hardware of physical set-up; the Peer, who takes part in the task or interaction, while suggesting goals/intentions for the supervisor to perform; 
and the Bystander, who may take part in the task or interaction through a subset of the available actions, while most likely not previously informed about which those are. These five roles were initially adapted from \ac{HCI} research, namely from Norman's \ac{HCI} Model~\cite{norman1986cognitive}. As such, they refer mostly to the role of the human within a technological system, whereas in this section we look for a classification to support the roles of robots in relation to humans within their interaction with each other.

Later, Goodrich et al.~\cite{goodrich2008human} built upon this list to propose a classification of roles that robots can assume in \ac{HRI}. In their list, it is not specified whether the role refers to a human or to a robot. 
Their proposed classification can be vague, as they take Scholtz's roles (for humans) and directly apply them to both robots and humans with no discussion provided. They also extended the list by adding two more roles, but these are defined only for robot. In the Mentor role, the robot is in a teaching or leadership role for the human; in the Informer role, the robot is not controlled by the human, but the human uses information coming from the robot, for example in a reconnaissance task.

The concept of robot roles was also addressed by Breazeal~\cite{breazeal2004social}, who proposed four interaction paradigms of \ac{HRI}. In these paradigms, the robot can either take the role of a Tool, directed at performing specific tasks, with various levels of autonomy; a Cyborg extension, in which it is physically merged with the human in a way that the person accepts it as an integral part of their body; an Avatar, through which the person can project themselves in order to communicate with another from far away; or a Sociable partner, as in classic science-fiction fantasy.

\begin{figure*}
    \centering
    \includegraphics[width=0.95\textwidth]{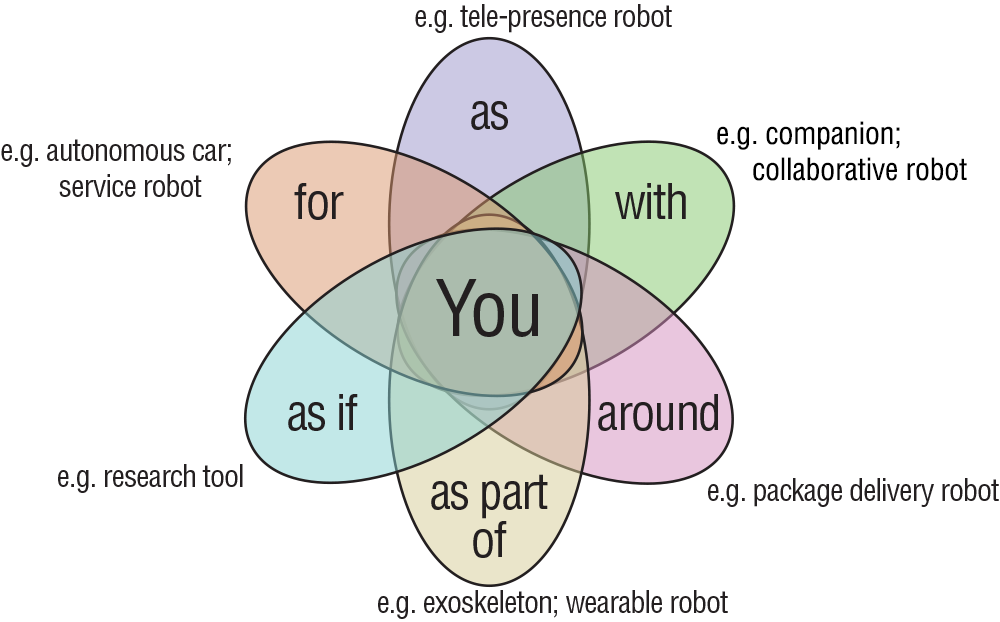}
    \caption{Our classification of relational roles of robots towards humans (represented as the ``you'').}
    \label{fig:roles}
\end{figure*}

Based on the many different proposed classifications, and of all the various interaction scenarios and applications found throughout literature and presented throughout this chapter, we have outlined our own classification for the role of robots within a relation with humans. Our classification attempts to merge the various dimensions of interaction while stepping away from explicit types of scenarios or applications. It does not necessarily add or propose new roles, but instead, redefines them from a relational perspective, placing emphasis on how the robot relates from a human's perspective, as depicted in Figure~\ref{fig:roles}.

In our classification for relational roles of robots, we view \ac{HRI} as including both \textbf{robot} and \textbf{you} (the human). 
As such, we consider the following roles that a \textit{robot} may have towards \textit{you}:
\begin{itemize}
\item A robot \textbf{``for you''} serves some utility on a given task.
This is the most traditional role of a tool or a servant, and is inspired by most previous classifications. 
Despite closely related with the concept of a tool, as proposed by other authors, we frame this role as a broader type of robotic tool, which can even include robots like autonomous cars.
\item A robot \textbf{``as you''} plays the role of a proxy, namely, but not limited to, tele-presence. However it does not necessarily imply interaction from far away as in Breazeal's classification~\cite{breazeal2004social}.
This type of role can exist even when inter-actors are co-located, as long as the robot is acting in place of another person who operates it (e.g. shared autonomy scenarios).
\item A robot \textbf{``with you''} is typically collaborative, with various levels of autonomy, including being part of a group with you. 
It is used in applications in which both the human and the robot act together, as a team, or towards common goals, and also includes robots for companionship.
The robot and human are not necessarily co-located, such as for example human-robot teams that have to communicate remotely.
\item A robot \textbf{``as if you''} emulates particular social or psychological traits found in humans. These robots are mainly used as social sciences research tools (see Section 2.3.8).
To date, robots have been used to examine, validate and refine theories of social and biological development, psychology, neurobiology, emotional and non-verbal communication, and social interaction.
\item A robot \textbf{``around you''}, shares a physical space and common resources with the human.
It differs from a \textit{robot with you} by the fact that it is necessarily co-located with the human, but not necessarily collaborating with them. 
These are typically called co-operating, co-present, or bystanders, as previously proposed in Scholzt's classification~\cite{scholtz2003theory}.
\item A robot \textbf{``as part of you''} extends the human body's capabilities. 
These robots typically have nonexistent or very limited autonomy, but provide humans with physical capabilities that they could not otherwise perform using their own biological body. Such robots can be used for pure embodiment extension (e.g. strength-enhancing exoskeletons), or for close-range \ac{HRI} collaboration, such as the robotic wearable forearm \cite{vatsal2018design} whose function is to serve as a supernumerary
third arm for shared workspace activities.
\end{itemize}

The list of relational roles that we present defines non-exclusive roles, meaning that for some particular applications, we may design and develop robots that take more than one of these roles, or take a different role when more than one human is involved in the interaction.
An example would be of a robot used in an office, which can be used \textit{for the users} to deliver mail and packages to different locations, while at the same time acting \textit{around the users} when navigating the office space. Another example would be an autonomous vehicle operating \textit{for} the passenger(s), but \textit{around} pedestrians and other human drivers.

\subsection{Autonomy and intelligence}
\label{subsec:autonomy}

Necessary aspects to consider when characterizing the behavior of social robots are those of autonomy and intelligence. Although related, these are two distinct concepts that are often inconsistently and confusingly used in existing literature~\cite{gunderson2004intelligence,gardner1996intelligence}. In particular, it is often assumed that a high level of robot autonomy implies both a high level of intelligence and of complexity. In reality, some fully autonomous systems can possess very low intelligence (e.g., a traditional manufacturing machine) or complexity (e.g., a simple self-operated mechanism). A better clarification of the concepts of autonomy and intelligence, and their relation, is needed, especially in the context of social robotics.

\subsubsection{Definitions (or lack thereof)}
The concepts of autonomy and intelligence are hard to define, and there does not seem to be unique accepted definitions~\cite{beer2014toward}. In particular, existing definitions in the literature seem to differ depending on the context of application, and the main field of focus of the author(s). Based on existing literature, we propose below extended working definitions of those two concepts in the context of social robotics.

\subsubsection{Autonomy}

It may seem somewhat paradoxical to talk about autonomy in the context of interactive robots, because traditionally fully autonomous robots are involved in minimal interactions with humans; in other words, reduced interaction with humans is a by-product of increased robot autonomy. For social robots however, this relation between amount of human interaction and robot autonomy is questioned. Highly autonomous social robots are expected to carry out more fluid, natural, and complex interactions, which does not make them any less autonomous. There exists a very large number of definitions of autonomy for general agents, however central to most existing definitions is the amount of \textit{control} the robot has over performing the task(s) it was designed to fulfill (or that it sets to itself), as emphasized by Beer et al.~\cite{beer2014toward}. For social robots, tasks may include well-defined goal states (e.g., assembling furniture) or more elusive ones (e.g., engaging in conversation). 

We claim that in addition to control, the concept of autonomy should also account for learning. Indeed, many learning paradigms include human-in-the-loop approaches, and we believe these should taken into account. These include active learning~\cite{chao2010transparent}, learning by demonstration~\cite{rybski2007interactive}, and corrective human feedback learning~\cite{mericcli2011task}, used within the context of interactions in applications involving human teachers such as learning-by-teaching educational scenarios~\cite{jacq2016building} or general collaborative scenarios~\cite{breazeal2004teaching}. As a result, we extend the definition from Beer et al.~\cite{beer2014toward} to make it applicable to social robots, and define autonomy of a social robot as follows:\\

\noindent\textit{\textbf{Autonomy ---} ``The extent to which a robot can operate in the tasks it was designed for (or that it creates for itself) without external intervention.''} \\

\noindent Note the use of the term \textit{intervention} as opposed to \textit{interaction}.

\subsubsection{Intelligence}

The is no real consensus on the definition of general intelligence~\cite{gardner1996intelligence}. In the context of robotics and \ac{AI}, intelligence is generally emphasized as related to problem solving~\cite{newell1972human}. For social robots, we propose the following extension of the definition of Gunderson et al.~\cite{gunderson2004intelligence}:\\

\noindent\textit{\textbf{Intelligence ---} ``The ability to determine behavior that will maximize the likelihood of goal satisfaction under dynamic and uncertain conditions, linked to the environment and the interaction with other (possibly human) agents.}''\\

Note that intelligence is also dependent on the difficulty of the goals to be achieved. Based on this definition, it can be seen that intelligence and autonomy are distinct concepts, but that, for a given task, intelligence creates a bound on achievable autonomy. In other words, the level of intelligence of a robot may prevent its ability to reach a given level of autonomy for fixed robot capabilities~\cite{gunderson2004intelligence}. A final important note concerning the design of social robots is that a robot's perceived intelligence~\cite{bartneck2009measurement} can be drastically different from its actual intelligence. As a result, minimizing the gap between the two is crucial for maintaining adequate expectations and appropriate levels of trust on the human side. Now that we have defined the concepts of autonomy and intelligence, we discuss approaches to quantify them.

\subsubsection{Quantifying autonomy and intelligence}
Unlike scales from the automation~\cite{endsley1999level} or tele-operation~\cite{sheridan1978human,huang2005autonomy,yanco2004classifying,goodrich2003seven} fields, and more recently with autonomous vehicles~\cite{sae2014automated}, all of which are based on the idea that more autonomy requires less HRI, some researchers have developed scales of autonomy that apply to social robots~\cite{beer2014toward,thrun2004toward,feil2007benchmarks,goodrich2008human}. These emphasize on the fact that autonomy has to be understood as a dynamic entity~\cite{goodrich2008human}. On the other hand, measuring robot intelligence has been the subject of some investigation, from both practical~\cite{adams2016athlon} and theoretical perspectives~\cite{bien2002machine}. Both autonomy and intelligence can be seen as belonging to a continuum, taking into account aspects of robot perception, cognition, execution, and learning~\cite{gunderson2004intelligence,yanco2004classifying}. As a result, autonomy is a dimension that one designs for, constrained by possible achievable levels of intelligence. As a general rule, the higher the autonomy and intelligence is, the higher the complexity of the system is.

\subsubsection*{The importance of dimensional thinking}

For a highly heterogeneous technology such as a social robot that involves a combination of hardware, software architecture, cognition mechanisms, intelligent hardware control, just to name a few, it is important to define dimensions about aspects such as autonomy and intelligence. The overall assessment of these aspects would then depend on a combination of assessments over individual dimensions. Researchers at IBM have proposed to define ``dimensions of (general artificial) intelligence'', as a way to define an updated version of the Turing test~\cite{turing2009computing}. Their list is more task-oriented, but can serve as a basis to think about general dimensions for both intelligence and autonomy. We propose the following dimensions of intelligence and autonomy, accounting for the socially interactive factor:

\begin{enumerate}
\item \textbf{Perception of environment-related and human-related factors ---} In order to engage in successful interactions, social robots need to be able to assess the dynamic state of the physical environment and of humans, to inform its decision making. On the human side, this includes estimating the human's physical parameters (pose, speed, motion, etc.), speech, and non-verbal social cues (gestures, gaze, prosody, facial expressions, etc.).
\item \textbf{Modeling of environment and human(s) ---}
In order to interpret robot perceptions, models of the environment and of humans are needed. For example, models of the humans can allow the robot to infer their intents, personality, emotional or affective states, and predict future human states or behavior. If models are parametrized to capture individual differences, then they can be a powerful tool to inform personalization and adaptation mechanisms in HRI~\cite{rossi2017user}.
\item \textbf{Planning actions to interact with environment and human(s) ---} Decision-making on a robot can be reduced to creating plans for robot actions that take into account the shape of the task, the goal, and the current state of the world, including the robot, the environment, and the human(s). A social robot needs to plan its motion, speech, and any other modality of social behavior it may be able to exhibit.
\item \textbf{Executing plans under physical and social constraints ---} The same way the environment poses physical constraints on how the robot interacts with it, culture and society impose social constraints on how interactions with a robot should take place \cite{lee2014culturally}. Robot decision-making should take human social norms into account while planning and executing generated plans~\cite{carlucci2015explicit}. Note that the execution of the plan may not be successful, hence the robot needs to account for all possible outcomes.
\item \textbf{Learning through interaction with the environment or humans ---} On top of the $4$ basic dimensions mentioned above, some robots may be endowed with learning capabilities, which allow them to improve with time, throughout their interactions with the environment or humans (including human-in-the-loop learning). Note that this dimension does not necessarily encompass machine learning as a general technique, as many offline machine learning methods would fall under the dimensions of perception and modeling.
\end{enumerate}

The dimensions above span most existing building blocks for the intelligence of a social robot. However, depending on their implementation and complexity, some robots may not include one or more of the above dimensions. Those dimensions are generally separated in the design and implementation of most robots, hence as a result, intelligence and autonomy on each dimension may be completely different. For example, some semi-autonomous robots include completely human-controlled perception~\cite{steinfeld2009oz}, or rely on human input for learning~\cite{chao2010transparent,rybski2007interactive,mericcli2011task} or verifying the suitability of robot plans~\cite{esteban2017build}.

As technology advances, higher amounts of robot intelligence will be achievable, unlocking new possible levels of autonomy for more complex tasks; however, the amount of autonomy of a system (within possible technological limits) will remain a design choice. As a design principle for future social robots, we advocate for the notion of symbiotic autonomy~\cite{veloso2015cobots, coradeschi2006symbiotic}, where both humans and robots can overcome their limitations and potentially learn from each other.

\subsection{Proximity}
\label{subsec:proximity}

Spatial features of the interaction may have a strong influence on the type of possible interactions and their perception by humans. 
In this section, we focus on the proximity of the interaction, i.e., the physical distance between the robot and the human. 
In particular, we consider $3$ general categories of interactions according to the proximity dimension: \textit{remote}, \textit{co-located}, and \textit{physical}.

\subsubsection{Remote \ac{HRI}}

Several applications in \ac{HRI} require the human and the robot to be in physically remote places. Tele-operation applications generally involve tasks or environments that are dangerous or inaccessible for humans, and historically represents one of the first involvements of humans with robots. 
In traditional tele-operation contexts, the human is treated as an operator, intervening to shape the behavior of one or more robots. 
Such types of \ac{HRI} scenarios have been extensively studied and a number of metrics have been developed for them~\cite{steinfeld2006common}. 
However, they are often excluded from the literature in social robotics~\cite{fong2002survey}.

More recent developments in the field of tele-operation gave rise to \textit{tele-presence} applications, which treat the robot as a physical proxy for the human~\cite{tsui2011exploring, kristoffersson2013review}, allowing the latter for example to be virtually present in tele-conferencing settings, or to visit remote places. As a result, as the robot is used to interact with humans in the remote environment, its design may include a strong focus on socially embodied aspects of the interaction beyond mere audio and video, such as distancing and gaze behavior~\cite{adalgeirsson2010mebot}. 

In all the previously cited literature, several notes are made regarding issues that are commonly faced, and should be addressed when developing social robots for tele-presence applications, such as concerns regarding privacy, a proper control interface for the pilot (including a map of the environment and the robot's surroundings), adaptability to people's height and stance (e.g., sitting, standing, behind a desk), robustness towards communication failures (e.g., loss of WiFi connection), and dynamic volume control. 

Finally, an important aspect of remote interaction is the translation of the operator's input into robot behaviors. Many interfaces have been developed for controlling tele-presence robots, including graphical and tangible interfaces~\cite{lazewatsky2011panorama}, but also virtual reality tools~\cite{nguyen2001virtual}, or brain-machine interfaces~\cite{tonin2011brain}.

\subsubsection{Co-located \ac{HRI}}
This category includes all interactions in which the robot and the human are located in a shared space and interact directly without explicit physical contact. This is the case for most existing social robotics scenarios.

Within these case we are most interested in mentioning the ones in which the robot has some form of locomotion ability (e.g., legged robot, aerial robots, wheeled robots), and also the ability to perceive and measure the distance to the human, in order to be able to actively control the distance between them. The social meaning of proximity in this context is referred to as proxemics, and constitutes an important part of non-verbal robot behavior~\cite{mumm2011human}.

Mead et al.~\cite{mead2016perceptual} have explored this topic by taking into account not only the psycho-physical and social aspects of proximity from the human's perspective, but also regarding the robot's needs. 
In terms of needs related to proximity, social robots may require or prefer certain distances to people in order for their sensors to work properly (e.g., vision, speech interaction).

Depending on the actual distance of the co-located robot, different modalities of communication may be more suitable. For example, robots in the private space may interact using speech or sound, and use touch screen for human input. However, robots at a greater distance but within line of sight, such as mobile robots, autonomous cars, or drones may use visual signals instead, such as expressive lights~\cite{baraka2018mobile,szafir2015communicating}.

\subsubsection{Physical \ac{HRI}}
Interactions happening in a shared space may involve an additional modality, namely physical contact between the human and the robot. Such interactions pertain to a blossoming subfield of HRI, commonly designated as Physical Human-Robot Interaction, or pHRI for short~\cite{Haddadin2016, billard2013roboskin, youssefi2015skinware}. From a hardware perspective, robots involved in pHRI are being designed with compliant joints (e.g., Baxter robot) for safety. Also, the design of robot outer shells is taking texture and feel into account~\cite{yohanan2009tool}. Moreover, novel paradigms for robot hardware are emerging with soft robotics~\cite{majidi2014soft}.

Examples of pHRI include physically assistive applications, where a robot has to be in physical contact with the person to execute its tasks, such as getting patients out of a chair~\cite{shomin2015sit}, or helping them feed~\cite{song2012novel} or dress themselves~\cite{kapusta2016data}. In industrial settings, physical proximity has also been shown, for some tasks, to improve the interaction and its perception by the workers~\cite{huber2017developing}.

On the other hand, physical contact may be used as a communication modality in itself, using a combination of touch, motion, pressure and/or vibration, known as haptic communication~\cite{miyashita2007haptic}. Such a communication modality is especially useful when others (e.g., visual) are not feasible. In particular, research has looked at how robots can communicate or guide people with visual impairments using physical contact. For example, Bonani et al.~\cite{bonani2018my} investigated the use of movement of a Baxter's arm that blind people held to complement verbal instructions in a playful assembly task. Additionally, mobile robots have been used to guide people in indoor environments using physical contact~\cite{kulyukin2004rfid,shomin2016navigation}.

Moreover, physical contact may possess a social component. This is the case when a robot behavior utilizing physical contact with a human is meant to induce or influence their behavior. For example, a mobile robot may use physical contact when navigating through a human crowded environment, inducing people to move away~\cite{shrestha2015using}. Also, affective robot behaviors involving contact, such as a hug or a handshake, have been shown to have an influence on the social behavior of the humans in their interaction with the robot (e.g., self-disclosure or general perception of the robot)~\cite{shiomi2017robot,avelino2018power}. Human-robot haptics have also been investigated by studying the role of physical contact in human-animal interactions~\cite{yohanan2012role}.\\

While the spatial features discussed in this section pertain to different fields of research, one would expect in future robotic technologies a range of interactions that would incorporate a combination of the three, according to the task and situation at hand.

\subsection{Temporal profile}
\label{subsec:temporal}
 
In this section, we look at time-related aspects of interactions with a social robot. Knowing the intended temporal profile of these interactions may have a strong impact on the design of such robots. We specifically discuss the \textit{timespan}, the \textit{duration}, and the \textit{frequency} of interactions. 

\subsubsection{Timespan}
Interactions with robots can be classified according to \textit{timespan}, meaning the period of time in which the human is exposed to the robot. We consider four timespan categories, namely \textit{short-term}, \textit{medium-term}, \textit{long-term}, and \textit{life-long}. There does not exist, in the \ac{HRI} literature, a quantitative way to establish the boundaries between these four categories, and as they may be context-dependent. Our aim is hence to provide a useful guideline for thinking about implications of such categories in the design of social robots, as well as their evaluation.

\begin{itemize}
\item \textbf{Short-term} interactions typically consist of a single or only a few consecutive interactions, e.g., a robot giving directions in a mall. Of special importance for these types of interactions are design factors that influence the first impression of the human towards the robot (e.g., appearance, size, motion ``at rest'', proxemics/approach behavior, initiation of the interaction). Usually very present in short-term interactions is the novelty effect, a fundamental characteristic of any innovation characterized by the newness or freshness of the innovation in the eyes of the adopter~\cite{wells2010effect}. It is a salient effect that plays a role in the adoption and use of novel media, characterized by higher initial achievements not because actual improvements occur, but due to the increased interest in technology~\cite{clark1983reconsidering}. This effect may help or harm the interaction depending on the its content and outcome, but it should be kept in mind in the design of robots for short-term use, also accounting for different expectations based on the users' demographics.
\item \textbf{Medium-term} interactions go beyond a single or a few interaction(s) but do not extend over a timespan long enough to be considered part of the long-term category. They typically span several days or weeks. An example is a robot used to teach children a module in their curriculum over a few weeks. During repeated interactions, the novelty effect may wear off after the first few interactions, resulting in potential loss of interest or changes in attitudes towards robots over time~\cite{gockley2005designing, kanda2004interactive}. When considering repeated interactions with the same robot, it is hence essential to take this dynamic aspect into account by incrementally incorporating novelty or change in the behavior of the robot as well as maintaining a sense of continuity across interactions~\cite{leite2013social,alves2019empathic}. This will help sustain engagement and satisfaction both within and across individual interactions.
 
\item \textbf{Long-term} interactions include prolonged interactions that go beyond the period needed for the novelty effect to fade~\cite{leite2013social}. An example is a personal robot operating in a home. Long-term interactions typically create a sense of predictability in the human to know they will encounter a subsequent interaction. Additionally, humans may start feeling a sense of attachment to the robot, and even develop relationships with it. In addition to the points mentioned for the medium-term category, it is crucial to consider how the robot can both personalize and adapt its interactions with the human. Personalization means that the robot will accommodate for inter-individual differences, usually focusing on static or semi-static features of the human such as personality, preferences, or abilities. Adaptation means that the robot accommodates for intra-individual changes, focusing on dynamic features of the human such as physical, psychological and emotional state, performance, or behavior. For surveys about personalization and adaptation in \ac{HRI}, please consult Rossi et al.~\cite{rossi2017user} and Ahmad et al.~\cite{ahmad2017systematic}. Personalization can also include a dynamic component; for example, an algorithm has been developed for an office robot to learn not only preferences of robot behaviors but also how to switch between them across interactions, according to personality traits of the human~\cite{baraka2015adaptive}.

\item \textbf{Life-long} interactions differ from long-term interactions by the fact that the human may go through large changes, for example, transitioning from childhood to adulthood, or progressively loosing some capabilities during old age. These types of interactions are much rarer with existing robots, but we do have examples that include robotic pets adopted in life-long timespans such as the AIBO or PARO robots. Another example is robots meant to accompany people until the end of their lives, such as robots assisting the elderly while gaining skills over time hence compensating for the decrease in their users' capabilities~\cite{georgiadis2016robotic}. In the future, the vision of robotic companions~\cite{dautenhahn2004robots} may include richer interactions including mutual learning and evolution, emotional support, and building deeper bidirectional relationships.
\end{itemize}

\subsubsection{Duration and frequency}
In addition to timespan, an important temporal aspect of the interaction is the average \textit{duration} of individual interactions. For example, a human can interact with a robot in short-term but prolonged interactions (e.g., in an educational context), or on the contrary in short interactions over a long timespan (e.g., office robot), or in other combinations and levels of the above. An important question to consider for longer durations is how to maintain engagement, especially with populations with a short attention span, such as children. For short durations, it is important to design for intuitiveness and efficiency of the interaction, in order to reduce the cognitive load or adaptation time of the human.

It is worth mentioning that duration is often imposed by the task itself, but may also be imposed by the human's willingness to end it. For example, the Roboceptionist~\cite{gockley2005designing} interacts with people in a building over large timespans. It was designed as a conversational chatbot, hence every person that interacts with it can initiate and end the interaction at any moment. The authors reported short interactions generally under $30$ seconds, and aimed at increasing this number by designing for long-term interactions with engagement in mind, using techniques from the field of drama.

In addition to timespan and duration, the \textit{frequency} of interactions plays a role in their human perception by humans, and in the resulting design considerations. The frequency of interactions with the same robot can vary from very occasional (e.g., robots in stores visited sporadically) to multiple times per day (e.g., workplace robots). For high frequencies, a lack of of incorporation of novelty, or at least variation in the robot's behavior, may result in fatigue and lack of engagement. Also, achieving continuity through memory is a particularly relevant factor~\cite{leite2013social}. Currently, the effect of frequency on the perception and effectiveness of interactions seems to be largely lacking in the \ac{HRI} literature. \\

This concludes our discussion of time-related aspects of the interaction, as well as the discussion of our framework as a whole. Before concluding this chapter, we provide a brief discussion of design approaches for social robots.

\section{Working within the social robot design space}
\label{sec:discussion}

The framework presented in this chapter outlined major dimensions of relevance to the understanding of existing social robots and the design of future ones. Moving forward, it effectively defines a \textit{design space} for social robots, where each of the aspects discussed will involve a set of design decisions. For example: What role should my robot play in relation to humans? What should it look like? What kind of social capabilities should it have? What level of autonomy is best fitted for the task(s) and should it be fixed? etc. Higher-level decisions in the design process also arise such as: Are the requirements feasible with current technology, or will it require developing new technology? What are the practical considerations associated with the ``theoretically best'' design, as well as the costs, and are they outweighed by the benefits?

The actual design process of social robots and their interactions with humans has benefited from a number of design approaches inspired by design practices from a variety of fields such as engineering, computer science, \ac{HCI}, and human factors. For example, some researchers in \ac{HRI} have looked at developing design patterns that can be reused without having to start from scratch every time~\cite{kahn2008design}. There generally exist three broad design approaches, each of which may be valid depending on the intended context and objectives: human-centered design, robot-centered design, and symbiotic design. We briefly discuss these approaches next.

\subsection{Robots as technology adapted to humans (human-centered design)}
Human-centered design (HCD) is the central paradigms of \ac{HCI}, and much of \ac{HRI} design as a result. It aims to involve the intended user population as part of most development stages, including identifying needs and requirements, brainstorming, conceptualizing, creating solutions, testing, and refining prototypes through an iterative design process~\cite{abras2004user}.

In the \ac{HRI} context, the main assumption is that humans have their own communication mechanisms and unconsciously expect robots to follow human social communication modalities, rules, conventions and protocols. Important aspects of the robot behavior and embodiment design that play a strong role in terms of the human's perception of the interaction include physical presence 
\cite{bainbridge2008effect}, size 
\cite{powers2007comparing}, embodiment~\cite{lee2006physically, wainer2007embodiment},
affective behaviors~\cite{leite2008emotional}, role expectations~\cite{dautenhahn2005robot}, just to cite a few. From an evaluation point of view, HCD relies a lot on subjective self-reports of users to measure their perceptions, and complement more objective measures such as task performance.

While many HCD approaches exist for social robots, one of particular interest is treating robots as expressive characters, i.e., robots with  \textit{the ability of expressing identity, emotion and intention during autonomous interaction with human users}~\cite{ribeiro2017}. Designing for expressivity can be achieved for example by bringing professional animators to work side by side with robotic and \ac{AI} programmers. The idea is to utilize concepts of animation developed over several decades~\cite{ThomasJohnston1995} and apply them to robotic platforms~\cite{breemen2004, takayama2011expressing, ribeiro2012, hoffman2012, gielniak2012, ribeiro2013}.

\subsection{Robots as goal-oriented technology (robot-centered design)}
Historically, robots were developed solely by engineers who carried little concern about the human beyond the interface. While the focus in \ac{HRI} has now shifted to a more human-centered approach as was discussed in the previous section, HCD as a general design paradigm has been criticized by many researchers who consider it to be harmful in some aspects~\cite{greenberg2008usability,norman2005human}. For example, it has been criticized for its focus on usability (how easy it is to use) as opposed to usefulness (what benefits it provides) and its focus on incremental contributions based on human input conditioned by current technologies, which prevents from pushing technological boundaries. Additionally, adapting the technology to the user may sometimes be more costly than having the user adapt to the technology.

As a result, there are cases where a more robot-centered approach may work best. Excessively adapting robots to humans may result in suboptimal performance, high cost of development, or unmatched expectations. It is important to recognize that in some cases, it may be better to ask the human to adapt to the robot (maybe through training) in order to achieve better performance on the long run. Humans have a much better ability to adapt than robots, and it is crucial to identify when robots should not adapt because it would be more efficient to ask or expect humans to do it~\cite{norman2005human}. In many cases, the robot may have needs that may incur an immediate cost on humans, but result in a better future performance. Examples include robots asking for help from humans when they face limitations~\cite{veloso2015cobots}, or teaching the robot to perform a certain task so that it can perform better in subsequent tasks. A robot-centered approach may also include the adaptation of our environments to make them suitable for robots. Examples include avoiding construction materials that are not compatible with the robot's sensors, interfacing the robot with building facilities (such as elevators), and so on.

 \subsection{Robots as symbiotic embodied agents (symbiotic design)}
Both approaches discussed above, whether human-centered or robot-centered, are valid approaches that one can use when designing social robots and their associated tasks. As a general design process for such robots, we advocate for the careful identification of strengths and weaknesses of each part and design for an increased symbiosis between the human(s) and the robot(s). One way to achieve this symbiosis is to adopt a holistic view that focuses on the overall system behavior, as a function of robot(s), human(s), and the environment~\cite{steinfeld2009oz}. For example, the CoBot robots are autonomous mobile robots~\cite{veloso2015cobots} servicing human users in a building, designed with the ability to utilize the presence of other humans in the environment (i.e., bypassers) to overcome their limitations. For instance, they ask for assistance in pressing the elevator button or putting objects in their basket since they do not have arms. This is an example of symbiotic autonomy where humans and robots service each other mutually in the same shared environment, and where both parties have to adapt to the other party's needs.

\section{Conclusion}
\label{sec:conclusion}

In this chapter, we have introduced a framework for characterizing social robots and their interactions with humans along principal dimensions reflecting important design considerations. In particular, we (1) presented a broad classification of robot appearances, (2) repositioned existing classifications of robot social capabilities, (3) discussed a cross-section of purposes and application areas, (4) provided a straightforward and broad classification of the robot's relational role, (5) clarified the related but distinct concepts of autonomy and intelligence, and discussed their quantification, (6) analyzed interactions according to their spatial features, and (7) looked at time-related aspects of the interactions. While this framework is aimed primarily at characterizing social robots by drawing from a large body of literature to illustrate the concepts discussed, it also serves as a useful guide to inform the design of future social robots. Towards this end, we briefly touched upon different design approaches, namely human-centered, robot-centered, and symbiotic.

Social robotics is a growing multidisciplinary field that brings closer aspects of human nature with aspects of robotic technology. The scope of what a social robot means, does, or serves, will be shaped by future developments in the field. In this journey towards creating interactive intelligent machines, we are hopeful that as they become more socially apt, they contribute to expanding, not reducing, the foundational aspects of our humanity.

\begin{acknowledgement}
We would first like to thank C\'{e}line Jost for inviting us to be part of this book project and for contributing to the initial stages of the manuscript. Additionally, this book chapter would have not been possible without the valuable comments and suggestions of Prof. Ana Paiva. We would also like to thank the participants and co-organizers of the \href{https://gaips.inesc-id.pt/hri-reading-group/}{HRI Reading Group at Instituto Superior T\'{e}cnico} for sparking many discussions that influenced the content of this chapter. We would finally like to acknowledge the Global Communication Center at CMU for their feedback on one of our drafts. K. Baraka acknowledges the CMU-Portugal INSIDE project grant CMUP-ERI/HCI/0051/2013 and Funda\c{c}\~{a}o para a Ci\^{e}ncia e a Tecnologia (FCT) grants with ref. SFRH/BD/128359/2017 and UID/CEC/50021/2019. P. Alves-Oliveira acknowledges a grant from FCT with ref. SFRH/BD/110223/2015. The views and conclusions in this document are those of the authors only.
\end{acknowledgement}

%
%
 \bibliographystyle{spmpsci}
 \bibliography{biblio}

\begin{thebibliography}{100}
\providecommand{\url}[1]{{#1}}
\providecommand{\urlprefix}{URL }
\expandafter\ifx\csname urlstyle\endcsname\relax
  \providecommand{\doi}[1]{DOI~\discretionary{}{}{}#1}\else
  \providecommand{\doi}{DOI~\discretionary{}{}{}\begingroup
  \urlstyle{rm}\Url}\fi

\bibitem{abras2004user}
Abras, C., Maloney-Krichmar, D., Preece, J.: User-centered design.
\newblock Bainbridge, W. Encyclopedia of Human-Computer Interaction. Thousand
  Oaks: Sage Publications \textbf{37}(4), 445--456 (2004)

\bibitem{adalgeirsson2010mebot}
Adalgeirsson, S.O., Breazeal, C.: Mebot: a robotic platform for socially
  embodied presence.
\newblock In: Proceedings of the 5th ACM/IEEE international conference on
  Human-robot interaction, pp. 15--22. IEEE Press (2010)

\bibitem{adams2016athlon}
Adams, S.S., Banavar, G., Campbell, M.: I-athlon: Towards a multidimensional
  turing test.
\newblock AI Magazine \textbf{37}(1), 78--84 (2016)

\bibitem{admoni2017social}
Admoni, H., Scassellati, B.: Social eye gaze in human-robot interaction: a
  review.
\newblock Journal of Human-Robot Interaction \textbf{6}(1), 25--63 (2017)

\bibitem{ahmad2017systematic}
Ahmad, M., Mubin, O., Orlando, J.: A systematic review of adaptivity in
  human-robot interaction.
\newblock Multimodal Technologies and Interaction \textbf{1}(3), 14 (2017)

\bibitem{alonso2014human}
Alonso-Mora, J., Siegwart, R., Beardsley, P.: Human-robot swarm interaction for
  entertainment: From animation display to gesture based control.
\newblock In: Proceedings of the 2014 ACM/IEEE international conference on
  Human-robot interaction, pp. 98--98. ACM (2014)

\bibitem{alves2017yolo}
Alves-Oliveira, P., Arriaga, P., Paiva, A., Hoffman, G.: Yolo, a robot for
  creativity: A co-design study with children.
\newblock In: Proceedings of the 2017 Conference on Interaction Design and
  Children, pp. 423--429. ACM (2017)

\bibitem{alves2018yolo}
Alves-Oliveira, P., Chandak, A., Cloutier, I., Kompella, P., Moegenburg, P.,
  Bastos~Pires, A.E.: Yolo-a robot that will make your creativity boom.
\newblock In: Companion of the 2018 ACM/IEEE International Conference on
  Human-Robot Interaction, pp. 335--336. ACM (2018)

\bibitem{alves2016psychological}
Alves-Oliveira, P., K{\"u}ster, D., Kappas, A., Paiva, A.: Psychological
  science in {HRI}: Striving for a more integrated field of research.
\newblock In: 2016 AAAI Fall Symposium Series (2016)

\bibitem{alves2019empathic}
Alves-Oliveira, P., Sequeira, P., Melo, F.S., Castellano, G., Paiva, A.:
  Empathic robot for group learning: A field study.
\newblock ACM Transactions on Human-Robot Interaction (THRI) \textbf{8}(1), 3
  (2019)

\bibitem{anderson2018greeting}
Anderson-Bashan, L., Megidish, B., Erel, H., Wald, I., Hoffman, G., Zuckerman,
  O., Grishko, A.: The greeting machine: An abstract robotic object for opening
  encounters.
\newblock In: 2018 27th IEEE International Symposium on Robot and Human
  Interactive Communication (RO-MAN), pp. 595--602. IEEE (2018)

\bibitem{augugliaro2014flight}
Augugliaro, F., Lupashin, S., Hamer, M., Male, C., Hehn, M., Mueller, M.W.,
  Willmann, J.S., Gramazio, F., Kohler, M., D'Andrea, R.: The flight assembled
  architecture installation: Cooperative construction with flying machines.
\newblock IEEE Control Systems \textbf{34}(4), 46--64 (2014)

\bibitem{avelino2018power}
Avelino, J., Moreno, P., Bernardino, A., Correia, F., Paiva, A., Catarino, J.,
  Ribeiro, P.: The power of a hand-shake in human-robot interactions.
\newblock In: 2018 IEEE/RSJ International Conference on Intelligent Robots and
  Systems (IROS), pp. 1864--1869. IEEE (2018)

\bibitem{bainbridge2008effect}
Bainbridge, W.A., Hart, J., Kim, E.S., Scassellati, B.: The effect of presence
  on human-robot interaction.
\newblock In: Robot and Human Interactive Communication, 2008. RO-MAN 2008. The
  17th IEEE International Symposium on, pp. 701--706. IEEE (2008)

\bibitem{balch2002robot}
Balch, T., Parker, L.E.: Robot teams: from diversity to polymorphism.
\newblock AK Peters/CRC Press (2002)

\bibitem{baraka2019optimization}
Baraka, K., Couto, M., Melo, F.S., Veloso, M.: An optimization approach for
  structured agent-based provider/receiver tasks.
\newblock In: Proceedings of the 18th International Conference on Autonomous
  Agents and MultiAgent Systems, pp. 95--103. International Foundation for
  Autonomous Agents and Multiagent Systems (2019)

\bibitem{baraka2019interactive}
Baraka, K., Melo, F.S., Veloso, M.: Interactive robots with model-based
  `autism-like' behaviors.
\newblock Paladyn, Journal of Behavioral Robotics \textbf{10}(1), 103--116
  (2019)

\bibitem{baraka2015adaptive}
Baraka, K., Veloso, M.: Adaptive interaction of persistent robots to user
  temporal preferences.
\newblock In: International Conference on Social Robotics, pp. 61--71. Springer
  (2015)

\bibitem{baraka2018mobile}
Baraka, K., Veloso, M.: Mobile service robot state revealing through expressive
  lights: Formalism, design, and evaluation.
\newblock International Journal of Social Robotics \textbf{10}(1), 65--92
  (2018)

\bibitem{bartneck2009measurement}
Bartneck, C., Kuli{\'c}, D., Croft, E., Zoghbi, S.: Measurement instruments for
  the anthropomorphism, animacy, likeability, perceived intelligence, and
  perceived safety of robots.
\newblock International journal of social robotics \textbf{1}(1), 71--81 (2009)

\bibitem{baxter2016characterising}
Baxter, P., Kennedy, J., Senft, E., Lemaignan, S., Belpaeme, T.: From
  characterising three years of {HRI} to methodology and reporting
  recommendations.
\newblock In: The Eleventh ACM/IEEE International Conference on Human Robot
  Interaction, pp. 391--398. IEEE Press (2016)

\bibitem{beer2014toward}
Beer, J.M., Fisk, A.D., Rogers, W.A.: Toward a framework for levels of robot
  autonomy in human-robot interaction.
\newblock Journal of Human-Robot Interaction \textbf{3}(2), 74--99 (2014)

\bibitem{belpaeme2018social}
Belpaeme, T., Kennedy, J., Ramachandran, A., Scassellati, B., Tanaka, F.:
  Social robots for education: A review.
\newblock Science Robotics \textbf{3}(21) (2018)

\bibitem{bethel2006auditory}
Bethel, C.L., Murphy, R.R.: Auditory and other non-verbal expressions of affect
  for robots.
\newblock In: 2006 AAAI Fall Symposium Series, Aurally Informed Performance:
  Integrating Machine Listening and Auditory Presentation in Robotic Systems,
  Washington, DC (2006)

\bibitem{bethel2008survey}
Bethel, C.L., Murphy, R.R.: Survey of non-facial/non-verbal affective
  expressions for appearance-constrained robots.
\newblock IEEE Transactions on Systems, Man, and Cybernetics, Part C
  (Applications and Reviews) \textbf{38}(1), 83--92 (2008)

\bibitem{bethel2009robots}
Bethel~Cindy, L.: Robots without faces: non-verbal social human-robot
  interaction.
\newblock Ph.D. thesis, dissertation/Ph.D.'s thesis]. University of South
  Florida (2009)

\bibitem{bien2002machine}
Bien, Z., Bang, W.C., Kim, D.Y., Han, J.S.: Machine intelligence quotient: its
  measurements and applications.
\newblock Fuzzy sets and systems \textbf{127}(1), 3--16 (2002)

\bibitem{billard2013roboskin}
Billard, A., Bonfiglio, A., Cannata, G., Cosseddu, P., Dahl, T., Dautenhahn,
  K., Mastrogiovanni, F., Metta, G., Natale, L., Robins, B., et~al.: The
  roboskin project: Challenges and results.
\newblock In: Romansy 19--Robot Design, Dynamics and Control, pp. 351--358.
  Springer (2013)

\bibitem{billard2007building}
Billard, A., Robins, B., Nadel, J., Dautenhahn, K.: Building {Robota}, a
  mini-humanoid robot for the rehabilitation of children with autism.
\newblock Assistive Technology \textbf{19}(1), 37--49 (2007)

\bibitem{biswas20161}
Biswas, J., Veloso, M.: The 1,000-km challenge: Insights and quantitative and
  qualitative results.
\newblock IEEE Intelligent Systems \textbf{31}(3), 86--96 (2016)

\bibitem{bonani2018my}
Bonani, M., Oliveira, R., Correia, F., Rodrigues, A., Guerreiro, T., Paiva, A.:
  What my eyes can't see, a robot can show me: Exploring the collaboration
  between blind people and robots.
\newblock In: Proceedings of the 20th International ACM SIGACCESS Conference on
  Computers and Accessibility, pp. 15--27. ACM (2018)

\bibitem{breazeal2003toward}
Breazeal, C.: Toward sociable robots.
\newblock Robotics and autonomous systems \textbf{42}(3-4), 167--175 (2003)

\bibitem{breazeal2004social}
Breazeal, C.: Social interactions in hri: the robot view.
\newblock IEEE Transactions on Systems, Man, and Cybernetics, Part C
  (Applications and Reviews) \textbf{34}(2), 181--186 (2004)

\bibitem{breazeal2004teaching}
Breazeal, C., Hoffman, G., Lockerd, A.: Teaching and working with robots as a
  collaboration.
\newblock In: Proceedings of the Third International Joint Conference on
  Autonomous Agents and Multiagent Systems-Volume 3, pp. 1030--1037. IEEE
  Computer Society (2004)

\bibitem{breazeal2004designing}
Breazeal, C.L.: Designing sociable robots.
\newblock MIT press (2004)

\bibitem{breemen2004}
Breemen, A.V.: {Animation engine for believable interactive user-interface
  robots}.
\newblock In: IEEE/RSJ International Conference on Intelligent Robots and
  Systems - IROS '04, vol.~3, pp. 2873--2878 (2004).
\newblock \doi{10.1109/IROS.2004.1389845}

\bibitem{broadbent2009acceptance}
Broadbent, E., Stafford, R., MacDonald, B.: Acceptance of healthcare robots for
  the older population: Review and future directions.
\newblock International journal of social robotics \textbf{1}(4), 319 (2009)

\bibitem{bruce2000robot}
Bruce, A., Knight, J., Listopad, S., Magerko, B., Nourbakhsh, I.R.: Robot
  improv: Using drama to create believable agents.
\newblock In: ICRA, p. 4003 (2000)

\bibitem{burgar2000development}
Burgar, C.G., Lum, P.S., Shor, P.C., Van~der Loos, H.M.: Development of robots
  for rehabilitation therapy: The {Palo Alto VA/Stanford} experience.
\newblock Journal of rehabilitation research and development \textbf{37}(6),
  663--674 (2000)

\bibitem{burton2013dolphins}
Burton, A.: Dolphins, dogs, and robot seals for the treatment of neurological
  disease.
\newblock The Lancet Neurology \textbf{12}(9), 851--852 (2013)

\bibitem{buschmann2009humanoid}
Buschmann, T., Lohmeier, S., Ulbrich, H.: Humanoid robot lola: Design and
  walking control.
\newblock Journal of physiology-Paris \textbf{103}(3-5), 141--148 (2009)

\bibitem{cabibihan2013robots}
Cabibihan, J.J., Javed, H., Ang, M., Aljunied, S.M.: Why robots? a survey on
  the roles and benefits of social robots in the therapy of children with
  autism.
\newblock International journal of social robotics \textbf{5}(4), 593--618
  (2013)

\bibitem{cannata2006design}
Cannata, G., D'Andrea, M., Maggiali, M.: Design of a humanoid robot eye: models
  and experiments.
\newblock In: 2006 6th IEEE-RAS International Conference on Humanoid Robots,
  pp. 151--156. IEEE (2006)

\bibitem{cappo2018online}
Cappo, E.A., Desai, A., Collins, M., Michael, N.: Online planning for
  human--multi-robot interactive theatrical performance.
\newblock Autonomous Robots pp. 1--16 (2018)

\bibitem{carlucci2015explicit}
Carlucci, F.M., Nardi, L., Iocchi, L., Nardi, D.: Explicit representation of
  social norms for social robots.
\newblock In: Intelligent Robots and Systems (IROS), 2015 IEEE/RSJ
  International Conference on, pp. 4191--4196. IEEE (2015)

\bibitem{chao2010transparent}
Chao, C., Cakmak, M., Thomaz, A.L.: Transparent active learning for robots.
\newblock In: Human-Robot Interaction (HRI), 2010 5th ACM/IEEE International
  Conference on, pp. 317--324. IEEE (2010)

\bibitem{chen2011survey}
Chen, G.D., Wang, C.Y., et~al.: A survey on storytelling with robots.
\newblock In: International Conference on Technologies for E-Learning and
  Digital Entertainment, pp. 450--456. Springer (2011)

\bibitem{clark1983reconsidering}
Clark, R.E.: Reconsidering research on learning from media.
\newblock Review of educational research \textbf{53}(4), 445--459 (1983)

\bibitem{colton2012computational}
Colton, S., Wiggins, G.A., et~al.: Computational creativity: The final
  frontier?
\newblock In: Ecai, vol. 2012, pp. 21--16. Montpelier (2012)

\bibitem{coradeschi2006symbiotic}
Coradeschi, S., Saffiotti, A.: Symbiotic robotic systems: Humans, robots, and
  smart environments.
\newblock IEEE Intelligent Systems \textbf{21}(3), 82--84 (2006)

\bibitem{dautenhahn2004robots}
Dautenhahn, K.: Robots we like to live with! a developmental perspective on a
  personalized, life-long robot companion.
\newblock In: Procs 13th IEEE Int Workshop on Robot and Human Interactive
  Communication, RO-MAN (2004)

\bibitem{dautenhahn2007socially}
Dautenhahn, K.: Socially intelligent robots: dimensions of human--robot
  interaction.
\newblock Philosophical transactions of the royal society B: Biological
  sciences \textbf{362}(1480), 679 (2007)

\bibitem{dautenhahn1999studying}
Dautenhahn, K., Billard, A.: Studying robot social cognition within a
  developmental psychology framework.
\newblock In: 3rd European Workshop on Advanced Mobile Robots, Eurobot 1999,
  pp. 187--194. IEEE (1999)

\bibitem{dautenhahn2005robot}
Dautenhahn, K., Woods, S., Kaouri, C., Walters, M.L., Koay, K.L., Werry, I.:
  What is a robot companion-friend, assistant or butler?
\newblock In: Intelligent Robots and Systems, 2005.(IROS 2005). 2005 IEEE/RSJ
  International Conference on, pp. 1192--1197. IEEE (2005)

\bibitem{disalvo2003seduction}
DiSalvo, C., Gemperle, F.: From seduction to fulfillment: the use of
  anthropomorphic form in design.
\newblock In: Proceedings of the 2003 international conference on Designing
  pleasurable products and interfaces, pp. 67--72. ACM (2003)

\bibitem{disalvo2002all}
DiSalvo, C.F., Gemperle, F., Forlizzi, J., Kiesler, S.: All robots are not
  created equal: the design and perception of humanoid robot heads.
\newblock In: Proceedings of the 4th conference on Designing interactive
  systems: processes, practices, methods, and techniques, pp. 321--326. ACM
  (2002)

\bibitem{dragan2013legibility}
Dragan, A.D., Lee, K.C., Srinivasa, S.S.: Legibility and predictability of
  robot motion.
\newblock In: Proceedings of the 8th ACM/IEEE international conference on
  Human-robot interaction, pp. 301--308. IEEE Press (2013)

\bibitem{ekman1992argument}
Ekman, P.: An argument for basic emotions.
\newblock Cognition \& emotion \textbf{6}(3-4), 169--200 (1992)

\bibitem{endsley1999level}
Endsley, M.R.: Level of automation effects on performance, situation awareness
  and workload in a dynamic control task.
\newblock Ergonomics \textbf{42}(3), 462--492 (1999)

\bibitem{epley2007seeing}
Epley, N., Waytz, A., Cacioppo, J.T.: On seeing human: a three-factor theory of
  anthropomorphism.
\newblock Psychological review \textbf{114}(4), 864 (2007)

\bibitem{esteban2017build}
Esteban, P.G., Baxter, P., Belpaeme, T., Billing, E., Cai, H., Cao, H.L.,
  Coeckelbergh, M., Costescu, C., David, D., De~Beir, A., et~al.: How to build
  a supervised autonomous system for robot-enhanced therapy for children with
  autism spectrum disorder.
\newblock Paladyn, Journal of Behavioral Robotics \textbf{8}(1), 18--38 (2017)

\bibitem{eyssel2017experimental}
Eyssel, F.: An experimental psychological perspective on social robotics.
\newblock Robotics and Autonomous Systems \textbf{87}, 363--371 (2017)

\bibitem{faber2009humanoid}
Faber, F., Bennewitz, M., Eppner, C., Gorog, A., Gonsior, C., Joho, D.,
  Schreiber, M., Behnke, S.: The humanoid museum tour guide {Robotinho}.
\newblock In: Robot and Human Interactive Communication, 2009. RO-MAN 2009. The
  18th IEEE International Symposium on, pp. 891--896. IEEE (2009)

\bibitem{fasola2011comparing}
Fasola, J., Mataric, M.: Comparing physical and virtual embodiment in a
  socially assistive robot exercise coach for the elderly.
\newblock Center for Robotics and Embedded Systems, Los Angeles, CA  (2011)

\bibitem{feil2005defining}
Feil-Seifer, D., Mataric, M.J.: Defining socially assistive robotics.
\newblock In: Rehabilitation Robotics, 2005. ICORR 2005. 9th International
  Conference on, pp. 465--468. IEEE (2005)

\bibitem{feil2007benchmarks}
Feil-Seifer, D., Skinner, K., Matari{\'c}, M.J.: Benchmarks for evaluating
  socially assistive robotics.
\newblock Interaction Studies \textbf{8}(3), 423--439 (2007)

\bibitem{fernandez2018may}
Fern{\'a}ndez-Llamas, C., Conde, M.A., Rodr{\'\i}guez-Lera, F.J.,
  Rodr{\'\i}guez-Sedano, F.J., Garc{\'\i}a, F.: May i teach you? students'
  behavior when lectured by robotic vs. human teachers.
\newblock Computers in Human Behavior \textbf{80}, 460--469 (2018)

\bibitem{fincannon2004evidence}
Fincannon, T., Barnes, L.E., Murphy, R.R., Riddle, D.L.: Evidence of the need
  for social intelligence in rescue robots.
\newblock In: Intelligent Robots and Systems, 2004 (IROS 2004). Proceedings.
  2004 IEEE/RSJ International Conference on, vol.~2, pp. 1089--1095. IEEE
  (2004)

\bibitem{fink2012anthropomorphism}
Fink, J.: Anthropomorphism and human likeness in the design of robots and
  human-robot interaction.
\newblock In: International Conference on Social Robotics, pp. 199--208.
  Springer (2012)

\bibitem{fong2002survey}
Fong, T., Nourbakhsh, I., Dautenhahn, K.: A survey of socially interactive
  robots: concepts, design and applications.
\newblock Tech. Rep. CMU-RI-TR-02-29, Robotics Institute, Carnegie Mellon
  University (2002)

\bibitem{forlizzi2004assistive}
Forlizzi, J., DiSalvo, C., Gemperle, F.: Assistive robotics and an ecology of
  elders living independently in their homes.
\newblock Human-Computer Interaction \textbf{19}(1), 25--59 (2004)

\bibitem{friedman2003hardware}
Friedman, B., Kahn~Jr, P.H., Hagman, J.: Hardware companions?: What online aibo
  discussion forums reveal about the human-robotic relationship.
\newblock In: Proceedings of the SIGCHI conference on Human factors in
  computing systems, pp. 273--280. ACM (2003)

\bibitem{frith2010social}
Frith, U., Frith, C.: The social brain: allowing humans to boldly go where no
  other species has been.
\newblock Philosophical Transactions of the Royal Society B: Biological
  Sciences \textbf{365}(1537), 165--176 (2010)

\bibitem{gardner1996intelligence}
Gardner, H., Kornhaber, M.L., Wake, W.K.: Intelligence: Multiple perspectives.
\newblock Harcourt Brace College Publishers (1996)

\bibitem{gates2007robot}
Gates, B.: A robot in every home.
\newblock Scientific American \textbf{296}(1), 58--65 (2007)

\bibitem{georgiadis2016robotic}
Georgiadis, D., Christophorou, C., Kleanthous, S., Andreou, P., Santos, L.,
  Christodoulou, E., Samaras, G.: A robotic cloud ecosystem for elderly care
  and ageing well: The growmeup approach.
\newblock In: XIV Mediterranean Conference on Medical and Biological
  Engineering and Computing 2016, pp. 919--924. Springer (2016)

\bibitem{gielniak2012}
Gielniak, M.J., Thomaz, A.L.: {Enhancing interaction through exaggerated motion
  synthesis}.
\newblock ACM/IEEE International Conference on Human-Robot Interaction - HRI
  '12 p. 375 (2012).
\newblock \doi{10.1145/2157689.2157813}

\bibitem{glas2016erica}
Glas, D.F., Minato, T., Ishi, C.T., Kawahara, T., Ishiguro, H.: Erica: The
  erato intelligent conversational android.
\newblock In: Robot and Human Interactive Communication (RO-MAN), 2016 25th
  IEEE International Symposium on, pp. 22--29. IEEE (2016)

\bibitem{gockley2005designing}
Gockley, R., Bruce, A., Forlizzi, J., Michalowski, M., Mundell, A., Rosenthal,
  S., Sellner, B., Simmons, R., Snipes, K., Schultz, A.C., et~al.: Designing
  robots for long-term social interaction.
\newblock In: Intelligent Robots and Systems, 2005 (IROS 2005). 2005 IEEE/RSJ
  International Conference on, pp. 1338--1343. IEEE (2005)

\bibitem{goetz2003matching}
Goetz, J., Kiesler, S., Powers, A.: Matching robot appearance and behavior to
  tasks to improve human-robot cooperation.
\newblock In: The 12th IEEE International Workshop on Robot and Human
  Interactive Communication, 2003. Proceedings. ROMAN 2003., pp. 55--60. Ieee
  (2003)

\bibitem{goodrich2003seven}
Goodrich, M.A., Olsen, D.R.: Seven principles of efficient human robot
  interaction.
\newblock In: SMC'03 Conference Proceedings. 2003 IEEE International Conference
  on Systems, Man and Cybernetics. Conference Theme-System Security and
  Assurance (Cat. No. 03CH37483), vol.~4, pp. 3942--3948. IEEE (2003)

\bibitem{goodrich2008human}
Goodrich, M.A., Schultz, A.C., et~al.: Human--robot interaction: a survey.
\newblock Foundations and Trends{\textregistered} in Human--Computer
  Interaction \textbf{1}(3), 203--275 (2008)

\bibitem{graf2009robust}
Graf, C., H{\"a}rtl, A., R{\"o}fer, T., Laue, T.: A robust closed-loop gait for
  the standard platform league humanoid.
\newblock In: Proceedings of the Fourth Workshop on Humanoid Soccer Robots in
  conjunction with the, pp. 30--37 (2009)

\bibitem{greenberg2008usability}
Greenberg, S., Buxton, B.: Usability evaluation considered harmful (some of the
  time).
\newblock In: Proceedings of the SIGCHI conference on Human factors in
  computing systems, pp. 111--120. ACM (2008)

\bibitem{gunderson2004intelligence}
Gunderson, J., Gunderson, L.: Intelligence $\neq$ autonomy $\neq$ capability.
\newblock Performance Metrics for Intelligent Systems, PERMIS  (2004)

\bibitem{Haddadin2016}
Haddadin, S., Croft, E.: Physical Human--Robot Interaction, pp. 1835--1874.
\newblock Springer International Publishing, Cham (2016).
\newblock \doi{10.1007/978-3-319-32552-1\_69}

\bibitem{ho2010modelling}
Ho, W.C., Dautenhahn, K., Lim, M.Y., Du~Casse, K.: Modelling human memory in
  robotic companions for personalisation and long-term adaptation in {HRI}.
\newblock In: BICA, pp. 64--71 (2010)

\bibitem{hochberg2012reach}
Hochberg, L.R., Bacher, D., Jarosiewicz, B., Masse, N.Y., Simeral, J.D., Vogel,
  J., Haddadin, S., Liu, J., Cash, S.S., van~der Smagt, P., et~al.: Reach and
  grasp by people with tetraplegia using a neurally controlled robotic arm.
\newblock Nature \textbf{485}(7398), 372 (2012)

\bibitem{hoffman2012}
Hoffman, G.: {Dumb robots, smart phones: A case study of music listening
  companionship}.
\newblock IEEE International Symposium on Robot and Human Interactive
  Communication - RO-MAN'12 pp. 358--363 (2012).
\newblock \doi{10.1109/ROMAN.2012.6343779}

\bibitem{hoffman2010effects}
Hoffman, G., Breazeal, C.: Effects of anticipatory perceptual simulation on
  practiced human-robot tasks.
\newblock Autonomous Robots \textbf{28}(4), 403--423 (2010)

\bibitem{homans1974social}
Homans, G.C.: Social behavior: Its elementary forms.
\newblock Harcourt Brace Jovanovich (1974)

\bibitem{huang2005autonomy}
Huang, H.M., Pavek, K., Albus, J., Messina, E.: Autonomy levels for unmanned
  systems (alfus) framework: An update.
\newblock In: Unmanned Ground Vehicle Technology VII, vol. 5804, pp. 439--449.
  International Society for Optics and Photonics (2005)

\bibitem{huber2017developing}
Huber, A., Weiss, A.: Developing human-robot interaction for an industry 4.0
  robot: How industry workers helped to improve remote-{HRI} to physical-{HRI}.
\newblock In: Proceedings of the Companion of the 2017 ACM/IEEE International
  Conference on Human-Robot Interaction, pp. 137--138. ACM (2017)

\bibitem{jacq2016building}
Jacq, A., Lemaignan, S., Garcia, F., Dillenbourg, P., Paiva, A.: Building
  successful long child-robot interactions in a learning context.
\newblock In: Human-Robot Interaction (HRI), 2016 11th ACM/IEEE International
  Conference on, pp. 239--246. IEEE (2016)

\bibitem{jensen2005robots}
Jensen, B., Tomatis, N., Mayor, L., Drygajlo, A., Siegwart, R.: Robots meet
  humans: Interaction in public spaces.
\newblock IEEE Transactions on Industrial Electronics \textbf{52}(6),
  1530--1546 (2005)

\bibitem{jordan1998human}
Jordan, P.W.: Human factors for pleasure in product use.
\newblock Applied ergonomics \textbf{29}(1), 25--33 (1998)

\bibitem{jorgensen2018interaction}
J{\o}rgensen, J.: Interaction with soft robotic tentacles.
\newblock In: Companion of the 2018 ACM/IEEE International Conference on
  Human-Robot Interaction, pp. 38--38. ACM (2018)

\bibitem{kahn2008design}
Kahn, P.H., Freier, N.G., Kanda, T., Ishiguro, H., Ruckert, J.H., Severson,
  R.L., Kane, S.K.: Design patterns for sociality in human-robot interaction.
\newblock In: Proceedings of the 3rd ACM/IEEE international conference on Human
  robot interaction, pp. 97--104. ACM (2008)

\bibitem{kanda2004interactive}
Kanda, T., Hirano, T., Eaton, D., Ishiguro, H.: Interactive robots as social
  partners and peer tutors for children: A field trial.
\newblock Human--Computer Interaction \textbf{19}(1-2), 61--84 (2004)

\bibitem{kanda2007two}
Kanda, T., Sato, R., Saiwaki, N., Ishiguro, H.: A two-month field trial in an
  elementary school for long-term human--robot interaction.
\newblock IEEE Transactions on robotics \textbf{23}(5), 962--971 (2007)

\bibitem{kapusta2016data}
Kapusta, A., Yu, W., Bhattacharjee, T., Liu, C.K., Turk, G., Kemp, C.C.:
  Data-driven haptic perception for robot-assisted dressing.
\newblock In: Robot and Human Interactive Communication (RO-MAN), 2016 25th
  IEEE International Symposium on, pp. 451--458. IEEE (2016)

\bibitem{kkedzierski2013emys}
K{\k{e}}dzierski, J., Muszy{\'n}ski, R., Zoll, C., Oleksy, A., Frontkiewicz,
  M.: Emys--emotive head of a social robot.
\newblock International Journal of Social Robotics \textbf{5}(2), 237--249
  (2013)

\bibitem{kennedy2015comparing}
Kennedy, J., Baxter, P., Belpaeme, T.: Comparing robot embodiments in a guided
  discovery learning interaction with children.
\newblock International Journal of Social Robotics \textbf{7}(2), 293--308
  (2015)

\bibitem{knight2011eight}
Knight, H.: Eight lessons learned about non-verbal interactions through robot
  theater.
\newblock In: International Conference on Social Robotics, pp. 42--51. Springer
  (2011)

\bibitem{kolling2016human}
Kolling, A., Walker, P., Chakraborty, N., Sycara, K., Lewis, M.: Human
  interaction with robot swarms: A survey.
\newblock IEEE Transactions on Human-Machine Systems \textbf{46}(1), 9--26
  (2016)

\bibitem{komatsu2012does}
Komatsu, T., Kurosawa, R., Yamada, S.: How does the difference between users'
  expectations and perceptions about a robotic agent affect their behavior?
\newblock International Journal of Social Robotics \textbf{4}(2), 109--116
  (2012)

\bibitem{kozima2009keepon}
Kozima, H., Michalowski, M.P., Nakagawa, C.: Keepon.
\newblock International Journal of Social Robotics \textbf{1}(1), 3--18 (2009)

\bibitem{Kozima08aplayful}
Kozima, H., Michalowski, M.P., Nakagawa, C., Kozima, H., Nakagawa, C., Kozima,
  H., Michalowski, M.P.: A playful robot for research, therapy, and
  entertainment (2008)

\bibitem{kristoffersson2013review}
Kristoffersson, A., Coradeschi, S., Loutfi, A.: A review of mobile robotic
  telepresence.
\newblock Advances in Human-Computer Interaction \textbf{2013}, 3 (2013)

\bibitem{kulyukin2004rfid}
Kulyukin, V., Gharpure, C., Nicholson, J., Pavithran, S.: {RFID} in
  robot-assisted indoor navigation for the visually impaired.
\newblock In: Intelligent Robots and Systems, 2004 (IROS 2004). Proceedings.
  2004 IEEE/RSJ International Conference on, vol.~2, pp. 1979--1984. IEEE
  (2004)

\bibitem{lazewatsky2011panorama}
Lazewatsky, D.A., Smart, W.D.: A panorama interface for telepresence robots.
\newblock In: Proceedings of the 6th international conference on Human-robot
  interaction, pp. 177--178. ACM (2011)

\bibitem{lee2014culturally}
Lee, H.R., Sabanovi{\'c}, S.: Culturally variable preferences for robot design
  and use in south korea, turkey, and the united states.
\newblock In: Proceedings of the 2014 ACM/IEEE international conference on
  Human-robot interaction, pp. 17--24. ACM (2014)

\bibitem{lee2006physically}
Lee, K.M., Jung, Y., Kim, J., Kim, S.R.: Are physically embodied social agents
  better than disembodied social agents?: The effects of physical embodiment,
  tactile interaction, and people's loneliness in human--robot interaction.
\newblock International Journal of Human-Computer Studies \textbf{64}(10),
  962--973 (2006)

\bibitem{leite2013social}
Leite, I., Martinho, C., Paiva, A.: Social robots for long-term interaction: a
  survey.
\newblock International Journal of Social Robotics \textbf{5}(2), 291--308
  (2013)

\bibitem{leite2008emotional}
Leite, I., Pereira, A., Martinho, C., Paiva, A.: Are emotional robots more fun
  to play with?
\newblock In: Robot and human interactive communication, 2008. RO-MAN 2008. The
  17th IEEE international symposium on, pp. 77--82. IEEE (2008)

\bibitem{levy2009love}
Levy, D.: Love and sex with robots: The evolution of human-robot relationships.
\newblock New York (2009)

\bibitem{li2009amoeba}
Li, B., Ma, S., Liu, J., Wang, M., Liu, T., Wang, Y.: Amoeba-i: a
  shape-shifting modular robot for urban search and rescue.
\newblock Advanced Robotics \textbf{23}(9), 1057--1083 (2009)

\bibitem{li2010cross}
Li, D., Rau, P.P., Li, Y.: A cross-cultural study: Effect of robot appearance
  and task.
\newblock International Journal of Social Robotics \textbf{2}(2), 175--186
  (2010)

\bibitem{li2015benefit}
Li, J.: The benefit of being physically present: A survey of experimental works
  comparing copresent robots, telepresent robots and virtual agents.
\newblock International Journal of Human-Computer Studies \textbf{77}, 23--37
  (2015)

\bibitem{liang2018simultaneous}
Liang, Y.S., Pellier, D., Fiorino, H., Pesty, S., Cakmak, M.: Simultaneous
  end-user programming of goals and actions for robotic shelf organization.
\newblock In: 2018 IEEE/RSJ International Conference on Intelligent Robots and
  Systems (IROS), pp. 6566--6573. IEEE (2018)

\bibitem{lin2011goqbot}
Lin, H.T., Leisk, G.G., Trimmer, B.: Goqbot: a caterpillar-inspired soft-bodied
  rolling robot.
\newblock Bioinspiration \& biomimetics \textbf{6}(2), 026,007 (2011)

\bibitem{liu2007modular}
Liu, H., Meusel, P., Seitz, N., Willberg, B., Hirzinger, G., Jin, M., Liu, Y.,
  Wei, R., Xie, Z.: The modular multisensory dlr-hit-hand.
\newblock Mechanism and Machine Theory \textbf{42}(5), 612--625 (2007)

\bibitem{loffler2018multimodal}
L{\"o}ffler, D., Schmidt, N., Tscharn, R.: Multimodal expression of artificial
  emotion in social robots using color, motion and sound.
\newblock In: Proceedings of the 2018 ACM/IEEE International Conference on
  Human-Robot Interaction, pp. 334--343. ACM (2018)

\bibitem{luber2012socially}
Luber, M., Spinello, L., Silva, J., Arras, K.O.: Socially-aware robot
  navigation: A learning approach.
\newblock In: Intelligent robots and systems (IROS), 2012 IEEE/RSJ
  international conference on, pp. 902--907. IEEE (2012)

\bibitem{lungarella2003beyond}
Lungarella, M., Metta, G.: Beyond gazing, pointing, and reaching: A survey of
  developmental robotics  (2003)

\bibitem{lungarella2003developmental}
Lungarella, M., Metta, G., Pfeifer, R., Sandini, G.: Developmental robotics: a
  survey.
\newblock Connection science \textbf{15}(4), 151--190 (2003)

\bibitem{madhani2009bringing}
Madhani, A.J.: Bringing physical characters to life.
\newblock In: Human-Robot Interaction (HRI), 2009 4th ACM/IEEE International
  Conference on, pp. 1--1. IEEE (2009)

\bibitem{majidi2014soft}
Majidi, C.: Soft robotics: a perspective--current trends and prospects for the
  future.
\newblock Soft Robotics \textbf{1}(1), 5--11 (2014)

\bibitem{mavridis2015review}
Mavridis, N.: A review of verbal and non-verbal human--robot interactive
  communication.
\newblock Robotics and Autonomous Systems \textbf{63}, 22--35 (2015)

\bibitem{mavrogiannis2019effects}
Mavrogiannis, C., Hutchinson, A.M., Macdonald, J., Alves-Oliveira, P., Knepper,
  R.A.: Effects of distinct robot navigation strategies on human behavior in a
  crowded environment.
\newblock In: 2019 14th ACM/IEEE International Conference on Human-Robot
  Interaction (HRI), pp. 421--430. IEEE (2019)

\bibitem{mead2016perceptual}
Mead, R., Matari{\'c}, M.J.: Perceptual models of human-robot proxemics.
\newblock In: Experimental Robotics, pp. 261--276. Springer (2016)

\bibitem{mericcli2011task}
Meri{\c{c}}li, {\c{C}}., Veloso, M., Ak{\i}n, H.L.: Task refinement for
  autonomous robots using complementary corrective human feedback.
\newblock International Journal of Advanced Robotic Systems \textbf{8}(2), 16
  (2011)

\bibitem{miyashita2007haptic}
Miyashita, T., Tajika, T., Ishiguro, H., Kogure, K., Hagita, N.: Haptic
  communication between humans and robots.
\newblock In: Robotics Research, pp. 525--536. Springer (2007)

\bibitem{mori1970uncanny}
Mori, M.: The uncanny valley.
\newblock Energy \textbf{7}(4), 33--35 (1970)

\bibitem{mumm2011designing}
Mumm, J., Mutlu, B.: Designing motivational agents: The role of praise, social
  comparison, and embodiment in computer feedback.
\newblock Computers in Human Behavior \textbf{27}(5), 1643--1650 (2011)

\bibitem{mumm2011human}
Mumm, J., Mutlu, B.: Human-robot proxemics: physical and psychological
  distancing in human-robot interaction.
\newblock In: Proceedings of the 6th international conference on Human-robot
  interaction, pp. 331--338. ACM (2011)

\bibitem{murphy2010human}
Murphy, R.R., Nomura, T., Billard, A., Burke, J.L.: Human--robot interaction.
\newblock IEEE robotics \& automation magazine \textbf{17}(2), 85--89 (2010)

\bibitem{nass2005improving}
Nass, C., Jonsson, I.M., Harris, H., Reaves, B., Endo, J., Brave, S., Takayama,
  L.: Improving automotive safety by pairing driver emotion and car voice
  emotion.
\newblock In: CHI'05 extended abstracts on Human factors in computing systems,
  pp. 1973--1976. ACM (2005)

\bibitem{nass1994computers}
Nass, C., Steuer, J., Tauber, E.R.: Computers are social actors.
\newblock In: Proceedings of the SIGCHI conference on Human factors in
  computing systems, pp. 72--78. ACM (1994)

\bibitem{newell1972human}
Newell, A., Simon, H.A., et~al.: Human problem solving, vol. 104.
\newblock Prentice-Hall Englewood Cliffs, NJ (1972)

\bibitem{nguyen2001virtual}
Nguyen, L.A., Bualat, M., Edwards, L.J., Flueckiger, L., Neveu, C., Schwehr,
  K., Wagner, M.D., Zbinden, E.: Virtual reality interfaces for visualization
  and control of remote vehicles.
\newblock Autonomous Robots \textbf{11}(1), 59--68 (2001)

\bibitem{norman1986cognitive}
Norman, D.A.: Cognitive engineering.
\newblock User centered system design \textbf{31}, 61 (1986)

\bibitem{norman2005human}
Norman, D.A.: Human-centered design considered harmful.
\newblock interactions \textbf{12}(4), 14--19 (2005)

\bibitem{pagliarini2009development}
Pagliarini, L., Lund, H.H.: The development of robot art.
\newblock Artificial Life and Robotics \textbf{13}(2), 401--405 (2009)

\bibitem{paiva2017empathy}
Paiva, A., Leite, I., Boukricha, H., Wachsmuth, I.: Empathy in virtual agents
  and robots: A survey.
\newblock ACM Transactions on Interactive Intelligent Systems (TiiS)
  \textbf{7}(3), 11 (2017)

\bibitem{pfeifer2001understanding}
Pfeifer, R., Scheier, C.: Understanding intelligence.
\newblock MIT press (2001)

\bibitem{pollack2002pearl}
Pollack, M.E., Brown, L., Colbry, D., Orosz, C., Peintner, B., Ramakrishnan,
  S., Engberg, S., Matthews, J.T., Dunbar-Jacob, J., McCarthy, C.E., et~al.:
  Pearl: A mobile robotic assistant for the elderly.
\newblock In: AAAI workshop on automation as eldercare, vol. 2002, pp. 85--91
  (2002)

\bibitem{pope2018stickman}
Pope, M.T., Christensen, S., Christensen, D., Simeonov, A., Imahara, G.,
  Niemeyer, G.: Stickman: Towards a human scale acrobatic robot.
\newblock In: 2018 IEEE International Conference on Robotics and Automation
  (ICRA), pp. 2134--2140. IEEE (2018)

\bibitem{powers2007comparing}
Powers, A., Kiesler, S., Fussell, S., Fussell, S., Torrey, C.: Comparing a
  computer agent with a humanoid robot.
\newblock In: Proceedings of the ACM/IEEE international conference on
  Human-robot interaction, pp. 145--152. ACM (2007)

\bibitem{ribeiro2013}
Ribeiro, T., Dooley, D., Paiva, A.: {Nutty Tracks: Symbolic Animation Pipeline
  for Expressive Robotics}.
\newblock ACM International Conference on Computer Graphics and Interactive
  Techniques Posters - SIGGRAPH '13 p. 4503 (2013)

\bibitem{ribeiro2012}
Ribeiro, T., Paiva, A.: {The Illusion of Robotic Life Principles and Practices
  of Animation for Robots}.
\newblock In: ACM/IEEE International Conference on Human-Robot Interaction -
  HRI '12, 1937, pp. 383--390 (2012)

\bibitem{ribeiro2017}
Ribeiro, T., Paiva, A.: Animating the {Adelino} robot with {ERIK}: The
  expressive robotics inverse kinematics.
\newblock In: Proceedings of the 19th ACM International Conference on
  Multimodal Interaction, ICMI 2017, pp. 388--396. ACM, New York, NY, USA
  (2017).
\newblock \doi{10.1145/3136755.3136791}.
\newblock \urlprefix\url{http://doi.acm.org/10.1145/3136755.3136791}

\bibitem{robert2018personality}
Robert, L.: Personality in the human robot interaction literature: A review and
  brief critique.
\newblock In: Robert, LP (2018). Personality in the Human Robot Interaction
  Literature: A Review and Brief Critique, Proceedings of the 24th Americas
  Conference on Information Systems, Aug, pp. 16--18 (2018)

\bibitem{rossi2017user}
Rossi, S., Ferland, F., Tapus, A.: User profiling and behavioral adaptation for
  {HRI}: A survey.
\newblock Pattern Recognition Letters \textbf{99}, 3--12 (2017)

\bibitem{rybski2007interactive}
Rybski, P.E., Yoon, K., Stolarz, J., Veloso, M.M.: Interactive robot task
  training through dialog and demonstration.
\newblock In: Proceedings of the ACM/IEEE international conference on
  Human-robot interaction, pp. 49--56. ACM (2007)

\bibitem{sae2014automated}
{SAE International}: Automated driving: levels of driving automation are
  defined in new {SAE} international standard {J3016} (2014)

\bibitem{sauppe2015social}
Saupp{\'e}, A., Mutlu, B.: The social impact of a robot co-worker in industrial
  settings.
\newblock In: Proceedings of the 33rd annual ACM conference on human factors in
  computing systems, pp. 3613--3622. ACM (2015)

\bibitem{scassellati2002theory}
Scassellati, B.: Theory of mind for a humanoid robot.
\newblock Autonomous Robots \textbf{12}(1), 13--24 (2002)

\bibitem{scassellati2003investigating}
Scassellati, B.: Investigating models of social development using a humanoid
  robot.
\newblock In: Neural Networks, 2003. Proceedings of the International Joint
  Conference on, vol.~4, pp. 2704--2709. IEEE (2003)

\bibitem{scassellati2012robots}
Scassellati, B., Admoni, H., Matari{\'c}, M.: Robots for use in autism
  research.
\newblock Annual review of biomedical engineering \textbf{14}, 275--294 (2012)

\bibitem{scholtz2003theory}
Scholtz, J.: Theory and evaluation of human robot interactions.
\newblock In: 36th Annual Hawaii International Conference on System Sciences,
  2003. Proceedings of the, pp. 10--pp. IEEE (2003)

\bibitem{schou2018skill}
Schou, C., Andersen, R.S., Chrysostomou, D., B{\o}gh, S., Madsen, O.:
  Skill-based instruction of collaborative robots in industrial settings.
\newblock Robotics and Computer-Integrated Manufacturing \textbf{53}, 72--80
  (2018)

\bibitem{seok2010peristaltic}
Seok, S., Onal, C.D., Wood, R., Rus, D., Kim, S.: Peristaltic locomotion with
  antagonistic actuators in soft robotics.
\newblock In: 2010 IEEE International Conference on Robotics and Automation,
  pp. 1228--1233. IEEE (2010)

\bibitem{shamsuddin2012initial}
Shamsuddin, S., Yussof, H., Ismail, L., Hanapiah, F.A., Mohamed, S., Piah,
  H.A., Zahari, N.I.: Initial response of autistic children in human-robot
  interaction therapy with humanoid robot nao.
\newblock In: Signal Processing and its Applications (CSPA), 2012 IEEE 8th
  International Colloquium on, pp. 188--193. IEEE (2012)

\bibitem{sharkey2012granny}
Sharkey, A., Sharkey, N.: Granny and the robots: ethical issues in robot care
  for the elderly.
\newblock Ethics and information technology \textbf{14}(1), 27--40 (2012)

\bibitem{sheridan1978human}
Sheridan, T.B., Verplank, W.L.: Human and computer control of undersea
  teleoperators.
\newblock Tech. rep., Massachussetts inst of tech Cambridge man-machine systems
  lab (1978)

\bibitem{shibata2004overview}
Shibata, T.: An overview of human interactive robots for psychological
  enrichment.
\newblock Proceedings of the IEEE \textbf{92}(11), 1749--1758 (2004)

\bibitem{shibata2011robot}
Shibata, T., Wada, K.: Robot therapy: a new approach for mental healthcare of
  the elderly--a mini-review.
\newblock Gerontology \textbf{57}(4), 378--386 (2011)

\bibitem{shidujaman2018roboquin}
Shidujaman, M., Zhang, S., Elder, R., Mi, H.: ``{RoboQuin}'': A mannequin robot
  with natural humanoid movements.
\newblock In: 2018 27th IEEE International Symposium on Robot and Human
  Interactive Communication (RO-MAN), pp. 1051--1056. IEEE (2018)

\bibitem{shiomi2009field}
Shiomi, M., Kanda, T., Glas, D.F., Satake, S., Ishiguro, H., Hagita, N.: Field
  trial of networked social robots in a shopping mall.
\newblock In: Intelligent Robots and Systems, 2009. IROS 2009. IEEE/RSJ
  International Conference on, pp. 2846--2853. IEEE (2009)

\bibitem{shiomi2017robot}
Shiomi, M., Nakata, A., Kanbara, M., Hagita, N.: A robot that encourages
  self-disclosure by hug.
\newblock In: International Conference on Social Robotics, pp. 324--333.
  Springer (2017)

\bibitem{shomin2016navigation}
Shomin, M.: Navigation and physical interaction with balancing robots.
\newblock Ph.D. thesis, Robotics Institute, Carnegie Mellon University (2016)

\bibitem{shomin2015sit}
Shomin, M., Forlizzi, J., Hollis, R.: Sit-to-stand assistance with a balancing
  mobile robot.
\newblock In: Robotics and Automation (ICRA), 2015 IEEE International
  Conference on, pp. 3795--3800. IEEE (2015)

\bibitem{short2014train}
Short, E., Swift-Spong, K., Greczek, J., Ramachandran, A., Litoiu, A., Grigore,
  E.C., Feil-Seifer, D., Shuster, S., Lee, J.J., Huang, S., et~al.: How to
  train your dragonbot: Socially assistive robots for teaching children about
  nutrition through play.
\newblock In: The 23rd IEEE international symposium on robot and human
  interactive communication, pp. 924--929. IEEE (2014)

\bibitem{shrestha2015using}
Shrestha, M.C., Nohisa, Y., Schmitz, A., Hayakawa, S., Uno, E., Yokoyama, Y.,
  Yanagawa, H., Or, K., Sugano, S.: Using contact-based inducement for
  efficient navigation in a congested environment.
\newblock In: Robot and Human Interactive Communication (RO-MAN), 2015 24th
  IEEE International Symposium on, pp. 456--461. IEEE (2015)

\bibitem{sirkin2015mechanical}
Sirkin, D., Mok, B., Yang, S., Ju, W.: Mechanical ottoman: how robotic
  furniture offers and withdraws support.
\newblock In: Proceedings of the Tenth Annual ACM/IEEE International Conference
  on Human-Robot Interaction, pp. 11--18. ACM (2015)

\bibitem{song2012novel}
Song, W.K., Kim, J.: Novel assistive robot for self-feeding.
\newblock In: Robotic Systems-Applications, Control and Programming. InTech
  (2012)

\bibitem{sparrow2002march}
Sparrow, R.: The march of the robot dogs.
\newblock Ethics and information Technology \textbf{4}(4), 305--318 (2002)

\bibitem{spence2014welcoming}
Spence, P.R., Westerman, D., Edwards, C., Edwards, A.: Welcoming our robot
  overlords: Initial expectations about interaction with a robot.
\newblock Communication Research Reports \textbf{31}(3), 272--280 (2014)

\bibitem{srinivasa2010herb}
Srinivasa, S.S., Ferguson, D., Helfrich, C.J., Berenson, D., Collet, A.,
  Diankov, R., Gallagher, G., Hollinger, G., Kuffner, J., Weghe, M.V.: Herb: a
  home exploring robotic butler.
\newblock Autonomous Robots \textbf{28}(1), 5 (2010)

\bibitem{srinivasan2012social}
Srinivasan, V., Henkel, Z., Murphy, R.: Social head gaze and proxemics scaling
  for an affective robot used in victim management.
\newblock In: 2012 IEEE International Symposium on Safety, Security, and Rescue
  Robotics (SSRR), pp. 1--2. IEEE (2012)

\bibitem{steinfeld2006common}
Steinfeld, A., Fong, T., Kaber, D., Lewis, M., Scholtz, J., Schultz, A.,
  Goodrich, M.: Common metrics for human-robot interaction.
\newblock In: Proceedings of the 1st ACM SIGCHI/SIGART conference on
  Human-robot interaction, pp. 33--40. ACM (2006)

\bibitem{steinfeld2009oz}
Steinfeld, A., Jenkins, O.C., Scassellati, B.: The oz of wizard: simulating the
  human for interaction research.
\newblock In: Proceedings of the 4th ACM/IEEE international conference on Human
  robot interaction, pp. 101--108. ACM (2009)

\bibitem{stiehl2009huggable}
Stiehl, W.D., Lee, J.K., Breazeal, C., Nalin, M., Morandi, A., Sanna, A.: The
  huggable: a platform for research in robotic companions for pediatric care.
\newblock In: Proceedings of the 8th International Conference on interaction
  Design and Children, pp. 317--320. ACM (2009)

\bibitem{stone2007intelligent}
Stone, P.: Intelligent autonomous robotics: A robot soccer case study.
\newblock Synthesis Lectures on Artificial Intelligence and Machine Learning
  \textbf{1}(1), 1--155 (2007)

\bibitem{sum2017robot}
Sun, A., Chao, C., Lim, H.A.: Robot and human dancing.
\newblock In: UNESCO CID 50th World Congress on Dance Research (2017)

\bibitem{szafir2015communicating}
Szafir, D., Mutlu, B., Fong, T.: Communicating directionality in flying robots.
\newblock In: Proceedings of the Tenth Annual ACM/IEEE International Conference
  on Human-Robot Interaction, pp. 19--26. ACM (2015)

\bibitem{takayama2011expressing}
Takayama, L., Dooley, D., Ju, W.: Expressing thought: improving robot
  readability with animation principles.
\newblock In: 2011 6th ACM/IEEE International Conference on Human-Robot
  Interaction (HRI), pp. 69--76. IEEE (2011)

\bibitem{tanaka2007socialization}
Tanaka, F., Cicourel, A., Movellan, J.R.: Socialization between toddlers and
  robots at an early childhood education center.
\newblock Proceedings of the National Academy of Sciences \textbf{104}(46),
  17,954--17,958 (2007)

\bibitem{ThomasJohnston1995}
Thomas, F., Johnston, O.: {The Illusion of Life: Disney Animation}.
\newblock Hyperion (1995)

\bibitem{thrun2004toward}
Thrun, S.: Toward a framework for human-robot interaction.
\newblock Human-Computer Interaction \textbf{19}(1), 9--24 (2004)

\bibitem{tonin2011brain}
Tonin, L., Carlson, T., Leeb, R., Mill{\'a}n, J.d.R.: Brain-controlled
  telepresence robot by motor-disabled people.
\newblock In: Engineering in Medicine and Biology Society, EMBC, 2011 Annual
  International Conference of the IEEE, pp. 4227--4230. IEEE (2011)

\bibitem{tsui2011exploring}
Tsui, K.M., Desai, M., Yanco, H.A., Uhlik, C.: Exploring use cases for
  telepresence robots.
\newblock In: Proceedings of the 6th international conference on Human-robot
  interaction, pp. 11--18. ACM (2011)

\bibitem{turing2009computing}
Turing, A.M.: Computing machinery and intelligence.
\newblock In: Parsing the Turing Test, pp. 23--65. Springer (2009)

\bibitem{vatsal2018design}
Vatsal, V., Hoffman, G.: Design and analysis of a wearable robotic forearm.
\newblock In: 2018 IEEE International Conference on Robotics and Automation
  (ICRA), pp. 1--8. IEEE (2018)

\bibitem{veloso2015cobots}
Veloso, M.M., Biswas, J., Coltin, B., Rosenthal, S.: Cobots: Robust symbiotic
  autonomous mobile service robots.
\newblock In: IJCAI, p. 4423. Citeseer (2015)

\bibitem{wainer2007embodiment}
Wainer, J., Feil-Seifer, D.J., Shell, D.A., Mataric, M.J.: Embodiment and
  human-robot interaction: A task-based perspective.
\newblock In: Robot and Human interactive Communication, 2007. RO-MAN 2007. The
  16th IEEE International Symposium on, pp. 872--877. IEEE (2007)

\bibitem{wei2013autonomous}
Wei, J., Dolan, J.M., Litkouhi, B.: Autonomous vehicle social behavior for
  highway entrance ramp management.
\newblock In: Intelligent Vehicles Symposium (IV), 2013 IEEE, pp. 201--207.
  IEEE (2013)

\bibitem{wells2010effect}
Wells, J.D., Campbell, D.E., Valacich, J.S., Featherman, M.: The effect of
  perceived novelty on the adoption of information technology innovations: a
  risk/reward perspective.
\newblock Decision Sciences \textbf{41}(4), 813--843 (2010)

\bibitem{williams2018thank}
Williams, T., Thames, D., Novakoff, J., Scheutz, M.: Thank you for sharing that
  interesting fact!: Effects of capability and context on indirect speech act
  use in task-based human-robot dialogue.
\newblock In: Proceedings of the 2018 ACM/IEEE International Conference on
  Human-Robot Interaction, pp. 298--306. ACM (2018)

\bibitem{yadollahi2018deictic}
Yadollahi, E., Johal, W., Paiva, A., Dillenbourg, P.: When deictic gestures in
  a robot can harm child-robot collaboration.
\newblock In: Proceedings of the 17th ACM Conference on Interaction Design and
  Children, CONF, pp. 195--206. ACM (2018)

\bibitem{yamokoski2019robonaut}
Yamokoski, J., Radford, N.: Robonaut, valkyrie, and nasa robots.
\newblock Humanoid Robotics: A Reference pp. 201--214 (2019)

\bibitem{yanco2004classifying}
Yanco, H.A., Drury, J.: Classifying human-robot interaction: an updated
  taxonomy.
\newblock In: systems, man and cybernetics, 2004 IEEE International Conference
  on, vol.~3, pp. 2841--2846. IEEE (2004)

\bibitem{yim2007modular}
Yim, M., Shen, W.M., Salemi, B., Rus, D., Moll, M., Lipson, H., Klavins, E.,
  Chirikjian, G.S.: Modular self-reconfigurable robot systems [grand challenges
  of robotics].
\newblock IEEE Robotics \& Automation Magazine \textbf{14}(1), 43--52 (2007)

\bibitem{yim2002modular}
Yim, M., Zhang, Y., Duff, D.: Modular robots.
\newblock IEEE Spectrum \textbf{39}(2), 30--34 (2002)

\bibitem{yohanan2009tool}
Yohanan, S., MacLean, K.E.: A tool to study affective touch.
\newblock In: CHI'09 Extended Abstracts on Human Factors in Computing Systems,
  pp. 4153--4158. ACM (2009)

\bibitem{yohanan2012role}
Yohanan, S., MacLean, K.E.: The role of affective touch in human-robot
  interaction: Human intent and expectations in touching the haptic creature.
\newblock International Journal of Social Robotics \textbf{4}(2), 163--180
  (2012)

\bibitem{youssefi2015skinware}
Youssefi, S., Denei, S., Mastrogiovanni, F., Cannata, G.: Skinware 2.0: a
  real-time middleware for robot skin.
\newblock SoftwareX \textbf{3}, 6--12 (2015)

\bibitem{zeglin2014herb}
Zeglin, G., Walsman, A., Herlant, L., Zheng, Z., Guo, Y., Koval, M.C., Lenzo,
  K., Tay, H.J., Velagapudi, P., Correll, K., et~al.: Herb's sure thing: A
  rapid drama system for rehearsing and performing live robot theater.
\newblock In: Advanced robotics and its social impacts (ARSO), 2014 IEEE
  Workshop on, pp. 129--136. IEEE (2014)

\end{thebibliography}

\end{document}